\documentclass[final,1p,times,authoryear]{elsarticle}
\usepackage{lineno}
\usepackage[utf8]{inputenc}
\usepackage{amsmath,amssymb}

\usepackage{graphicx}
\usepackage{booktabs}
\usepackage{longtable}
\usepackage{tabularx}
\usepackage{ragged2e}
\newcolumntype{L}[1]{>{\RaggedRight\arraybackslash}p{#1}}
\newcolumntype{C}[1]{>{\Centering\arraybackslash}p{#1}}
\newcolumntype{R}[1]{>{\RaggedLeft\arraybackslash}p{#1}}

\usepackage{tikz}
\usepackage{neuralnetwork}
\usepackage{array}
\usepackage{colortbl}
\usepackage{siunitx}
\usepackage{multirow}
\usepackage[normalem]{ulem}
\useunder{\uline}{\ul}{}

\usepackage{natbib}            
\usepackage{microtype}
\usepackage[hyphens]{url}
\usepackage[hidelinks]{hyperref}
\usepackage{doi}
\usepackage[acronym]{glossaries}
\makeglossaries

\newacronym{dl}{DL}{Deep Learning}
\newacronym{1d-cnn}{1D-CNN}{One-Dimensional Convolutional Neural Networks}
\newacronym{2d-cnn}{2D-CNN}{Two-Dimensional Convolutional Neural Networks}

\newacronym{auc-roc}{AUC-ROC}{Area Under the Receiver Operating Characteristic Curve}
\newacronym{tpr}{TPR}{True Positive Rate}
\newacronym{fpr}{FPR}{False Positive Rate}

\newacronym{ml}{ML}{Machine Learning}
\newacronym{Hom_amer}{Homarus Americanus}{American lobster}

\newacronym{svm}{SVM}{Support Vector Machine}
\newacronym{mlp}{MLP}{Multi Layer Perceptron}
\newacronym{mfcc}{MFCC}{Mel Frequency Cepstral Coefficients}
\newacronym{knn}{KNN}{k-Nearest Neighbors}

\newacronym{1d-dcnn}{1D-DCNN}{One-Dimensional Deep Convolutional Neural Networks}
\newacronym{rf}{RF}{Random Forest}
\newacronym{xgboost}{XGBOOST}{Extreme Gradient Boosting}
\newacronym{nb}{NB}{Naive Bayes}
\newacronym{ai}{AI}{Artificial Intelligence}

\newacronym{Hom_euro}{Homarus Gammarus}{European lobster}
\newacronym{pan_argus}{Panulirus argus}{Caribbean spiny lobster}
\newacronym{pan_elephas}{Panulirus elephas}{European spiny lobster}
\newacronym{pan_ornatus}{Panulirus ornatus}{Tropical spiny lobster}

\newacronym{pam}{PAM}{Passive Acoustic Monitoring}

\newacronym{pca}{PCA}{Principal Component Analysis}

\newacronym{it}{IT}{Inference Time}
\newacronym{hl}{HL}{Hidden Layer}
\newacronym{lr}{LR}{Linear Regression}
\newacronym{tev}{TEV}{Total Explained Variance}

\newacronym{cnn}{CNN}{Convolutional Neural Network}
\newacronym{dbn}{DBN}{Deep Belief Network}
\newacronym{wndchrm}{WNDCHRM}{an open source utility tool for biological image analysis}
\newacronym{fft}{FFT}{Fast Fourier Transform}

\newacronym{pargus}{Panulirus argus}{caribbean spiny lobster}
\newacronym{pinterruptus}{Panulirus interruptus}{California spiny lobster}
\newacronym{Pelephas}{Palinurus elephas}{Red spiny lobster}
\newacronym{Jedwardsii}{Jasus edwardsii}{Australian spiny lobster}
\newacronym{fp}{FP}{False Positive}
\newacronym{fn}{FN}{False Negative}
\newacronym{tn}{TN}{True Negative}
\newacronym{tp}{TP}{True Positive}

\newacronym{eu}{EU}{European Union}
\newacronym{uiot}{UIOT}{Underwater Internet of Things}

\usepackage{tikz}
\usetikzlibrary{shapes.geometric, arrows.meta, positioning, calc} 
\usepackage{microtype} 

\usepackage{geometry}
\usepackage{lipsum}

\journal{Ecological Informatics}

\begin{document}
\begin{frontmatter}



\title{Sex and age determination in European lobsters using AI-Enhanced bioacoustics}


\author[first]{Feliciano Pedro Francisco Domingos}
\author[first]{Isibor Kennedy Ihianle}
\author[first]{Omprakash Kaiwartya}
\author[first]{Ahmad Lotfi}
\author[third]{Nicola Khan}
\author[second]{Nicholas Beaudreau}
\author[second]{Amaya Albalat}
\author[first]{Pedro Machado}

\affiliation[first]{organisation={Nottingham Trent University}, Department of Computer Science
            addressline={Erasmus Darwin Building}, 
            city={Clifton},
            postcode={NG11 8NS}, 
            state={Nottingham},
            country={United Kingdom}}

\affiliation[second]{organisation={University of Stirling},
            addressline={Pathfoot Building},
             city={Stirling},
             postcode={FK9 4LA},
            state={Scotland},
            country={Scotland, United Kingdom}}
\affiliation[third]{organisation={Ace Aquatec},
            addressline={1 Water's Edge, Camperdown Street},
             city={Dundee},
             postcode={DD1 3HY},
            state={Scotland},
            country={Scotland, United Kingdom}}

\begin{abstract}
Monitoring aquatic species presents considerable challenges due to their elusive nature and complex habitats. Consequently, the development and application of innovative, non-invasive approaches, such as \gls*{pam}, are paramount for effective ecological assessment and management. This study addresses these challenges by focusing on the \gls*{Hom_euro}, a key representative species of rocky benthic environments that underpins valuable local fisheries and aquaculture ventures. Comprehensive understanding of lobster habitats, welfare, reproduction, sex, and age is critical for robust aquaculture management, ecological research, conservation strategies and sustainable fisheries. While bioacoustic emissions have been successfully employed to classify various aquatic species using \gls{ai} models, such as fish, this research specifically leverages lobster bioacoustics to classify \gls*{Hom_euro} by age group (juvenile and adult) and sex (male and female). Despite lacking vocal cords, different types of lobsters produce a variety of characteristic sounds, such as stridulation characterised by the \gls{pan_elephas} and \gls{pan_argus}, buzzing or carapace vibrations expressed by the \gls*{Hom_euro} and \gls{Hom_amer}, rattling sounds marked by the \gls*{pan_ornatus} and clicking or snapping sounds. These varied lobster sounds are amenable to classification using advanced computational \gls{ai} models. The dataset collection was carried at Johnshaven in Scotland at the local Lobster Shop of Murray McBay and Company. Hydrophones were used and installed underwater in concrete tanks to record the lobster sounds. This investigation explores the efficacy of \gls*{dl} models (specifically \gls*{1d-cnn} and \gls*{1d-dcnn} with varying hidden layers) and six commonly employed \gls*{ml} models (\gls*{svm}, \gls*{knn}, \gls*{nb}, \gls*{rf}, \gls*{xgboost}, and \gls*{mlp} in classifying buzzing sounds or carapace vibrations of \gls*{Hom_euro} for age and sex classification. The \gls*{mfcc} served as features for all models across four distinct datasets. \gls*{mfcc} has been reported to extract reliable features from audio signals. The majority of models achieved classification accuracies exceeding 97\% for adult versus juvenile differentiation, with the exception of \gls*{nb}, which attained 91.31\%. For sex classification, all models except \gls*{nb} surpassed 93.23\% accuracy. These compelling results underscore the substantial potential of supervised \gls*{ml} and \gls*{dl} models to extract age- and sex-related features from lobster sounds. Ultimately, this research offers a promising non-invasive approach for lobster conservation, detection, and management within aquaculture and fisheries, thereby enabling real-world edge computing applications for \gls*{pam} of underwater species. \end{abstract}

\begin{keyword}
Lobster bioacoustics\sep \gls{ai} \sep \gls{ml} \sep \gls{dl} \sep age and sex classification \sep aquaculture management \sep \gls*{pam}.

\end{keyword}

\end{frontmatter}

\section{Introduction}\label{introduction}
Effective lobster aquaculture management, ecological research, and conservation strategies require a detailed understanding of habitat use, welfare, reproductive conditions, and the accurate identification of sex and age \cite{barroso2023applications, looby2023fishsounds}. Bioacoustics has emerged as a powerful tool to detect and monitor behavioural and ecological traits, with demonstrated success across avian and aquatic taxa, including fish and invertebrates \cite{BARDELI20101524, noda_automa, malfante2018automatic}. Lobsters produce diverse acoustic signals depending on species, including stridulation (\gls{pan_elephas}), buzzing or carapace vibrations (\gls*{Hom_amer}, \gls*{Hom_euro}), rattling (\gls{pan_ornatus}), and clicking or snapping sounds \cite{patek2002squeaking}. These emissions can be systematically classified using artificial intelligence (\gls*{ai}) models \cite{patricia_aggres2014}. The present study focuses on classifying buzzing sounds (carapace vibrations) produced by \gls{Hom_euro} to distinguish individuals by age class (adult vs.~juvenile) and sex (male vs.~female).\par

Sound production in lobsters arises from mechanical actions: \gls{pan_elephas} and \gls{pan_ornatus} generate stridulation by rubbing or contracting body parts, whereas \gls{Hom_euro} emit vibrations of the carapace in the 80--250~Hz range \cite{patek2002squeaking, staaterman2010, kerkhofs2024hum}. Such bioacoustic emissions are vital for aquatic life, serving functions including predatory avoidance, defence, aggression, courtship, and mating \cite{domingos2024underwater, looby2023fishsounds, barroso2023applications, weilgart2018impact}. More than 900 species of fish and invertebrates are known to produce active sounds, reflecting the centrality of acoustic communication in marine ecosystems \cite{jezequel2020acoustic}. For lobster aquaculture and fisheries, reliable and efficient methods for sex and age determination are critical to optimise yields, welfare, and sustainability \cite{mcgarvey2015comparing}. Traditional manual methods are time-consuming, labour-intensive, and prone to human error, motivating exploration of \gls*{dl} approaches such as \gls*{1d-cnn}, \gls*{1d-dcnn}, alongside established \gls{ml} models, for classification of \gls{Hom_euro} acoustic signals.

\begin{figure}[h]
\centering
\includegraphics[width=0.9\linewidth]{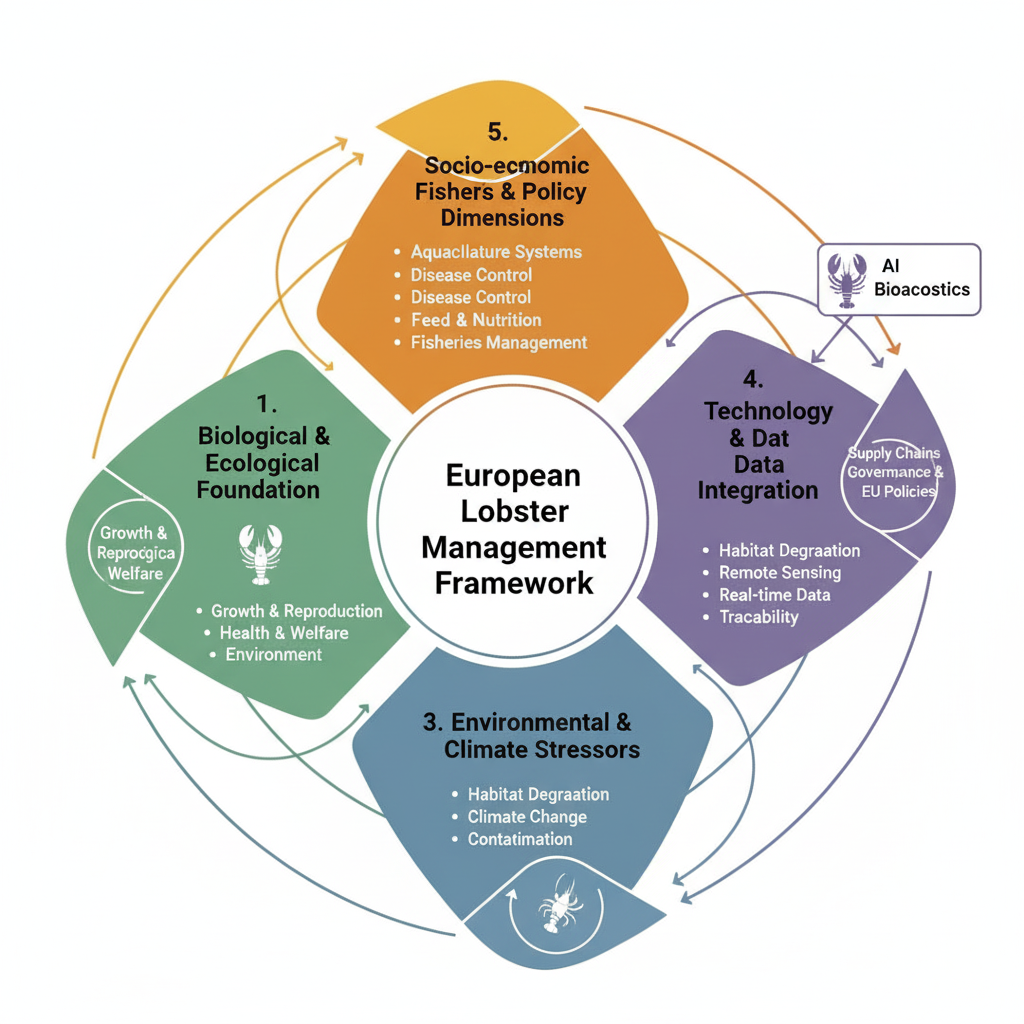}
\caption{Conceptual diagram of lobster management context showing how AI-enhanced bioacoustics for sex and age determination fits within broader ecological and policy considerations.}
\label{fig:lobster_context}
\end{figure}

To situate our study within broader management contexts, it is essential to note that effective lobster management extends beyond sex and age identification. Population dynamics are shaped by distribution patterns, life-history traits, spawning behaviours, and environmental pressures, including climate change \citep{Araujo2018ClimateLobsters,Phillips2015LobstersChangingTimes,Radhakrishnan2020EcologyDistributionLobsters}. Emerging threats, such as contamination by microplastics and heavy metals, further impact lobster health and seafood safety \citep{Hwang2019MicroplasticsSeafood,Yadav2024MPHeavyMetalsReview}. Figure~\ref{fig:lobster_context} illustrates how bioacoustics-based sex and age classification integrates within this wider ecological and policy framework.

In addition, regulatory frameworks such as \gls{eu} fisheries policies are central to stock sustainability, though they face ongoing scrutiny regarding ecological effectiveness and socio-economic trade-offs \citep{EU2025CFPOverview,Consilium2025CFPGoals}. Thus, management decisions are embedded in governance as much as in biology.

Our contribution addresses this interdisciplinary challenge by applying \gls{ai} models to classify lobster bioacoustics. Recent advances in \gls{ml} and \gls{dl} have demonstrated strong potential for detecting and classifying acoustic signatures of fish and crustaceans, enabling improved resource monitoring \citep{Niu2023MLUnderwaterAcousticsReview,Bharath2025UnderwaterSoundReview,Rountree2023FishSoundMLReview}. We argue that AI-enhanced bioacoustic monitoring offers a scalable, non-invasive method for lobster aquaculture and conservation, complementing ecological, socio-economic, and policy dimensions.

The main contributions of this article are as follows: (i) classification of lobster bioacoustics to determine age and sex; (ii) comparative evaluation of traditional \gls{ml} and modern \gls{dl} models, including four \gls*{1d-cnn} and four \gls*{1d-dcnn} architectures; (iii) analysis of model performance metrics to identify the most effective approaches; and (iv) a critical synthesis of related works on lobster acoustics and AI classification.

\section{Related Works} \label{sec:rework}

\subsection{Sound Production and Detection in Lobsters}
The understanding of how marine invertebrates produce and perceive sound remains limited \cite{taylor2010crustacean, jezequel2020acoustic}. In both \gls*{Hom_amer} and \gls*{Hom_euro}, vibrations and noise are generated by opposing remotor and promotor muscle contractions beneath the second antenna \cite{kerkhofs2024hum}. The abdominal vibrations of \gls*{Hom_amer} typically exhibit a base frequency of approximately 180~Hz, consistent with earlier reports \cite{staaterman2010}. For sound detection, external cuticular hairs appear to play a key role, while evidence suggests that crustacean statocysts are not primary auditory organs \cite{jezequel2021sound}. Spiny lobsters (\textit{Panulirus} spp.) also produce sounds through stick–slip mechanisms, where antennal bases rub against the antennular plate \cite{patek2002squeaking}. 

\subsection{Behavioural Context of Acoustic Emissions}
Lobster acoustic emissions are closely tied to behavioural states \cite{jezequel2018sound, looby2023fishsounds, domingos2024underwater}. Stressed \gls*{Hom_euro} individuals vibrate their carapace, producing low-frequency ``buzzing'' sounds similar to \gls*{Hom_amer} \cite{jezequel2018sound, jezequel2021sound, jezequel2020acoustic}. During feeding, lobsters emit broadband, transient ``rattles,'' comparable to those of tropical spiny lobsters (\textit{Palinurus longipes}, \textit{P.~argus}) \cite{jezequel2018sound}. Spiny lobsters also stridulate in anti-predatory contexts \cite{Bouwma01032009, kikuchi2015passive}. Male \gls*{Hom_euro} produce low-frequency buzzing during agonistic encounters, though tank-based hydrophones capture only $\sim$15\% of such signals due to attenuation \cite{jezequel2020acoustic}. Accelerometer-based detection has been employed to overcome these challenges.

\subsection{Propagation and Environmental Considerations}
Under natural low-noise conditions, spiny lobster sounds propagate over several kilometers, with the largest individuals exceeding 3~km \cite{jezequel2020spiny}. However, tank environments introduce reverberation and attenuation, which complicates analysis. Artificial sound source experiments demonstrated the importance of characterising tank acoustics before interpreting lobster behaviour from such recordings \cite{jezequel2018sound}. These environmental factors highlight both the potential and limitations of laboratory-based bioacoustic studies.

\subsection{Applications of AI in Marine Bioacoustics}
Beyond ecological studies, machine learning (ML) and deep learning (DL) approaches are increasingly used for acoustic classification. Discriminative ML models, such as Random Forests (RF) and Support Vector Machines (SVM), have achieved high accuracy in fish sound classification (up to 96.9\%) \cite{malfante2018automatic}. ML techniques are also being explored for undersea communication and environmental monitoring, offering noise reduction, adaptive modulation, and robust data handling \cite{domingos2024underwater, islam2025role}. 

DL architectures, including Convolutional Neural Networks (CNNs) and Deep Belief Networks (DBNs), have shown superior performance in underwater acoustic target classification, achieving accuracies above 94\% \cite{yue2017cnn_dbn}. While MFCCs remain standard features for ML models \cite{davis1980comparison, slaney1998auditorytoolbox}, CNNs and DBNs can directly process raw signals or FFT-based features, reducing reliance on handcrafted features. For lobster acoustics specifically, \gls*{1d-cnn} and dilated CNNs (\gls*{1d-dcnn}) offer the ability to capture temporal dynamics directly from raw or minimally processed signals \cite{huang2019signal}. 

\subsection{Towards Predictive Modelling in Lobster Bioacoustics}
Recent advances emphasise predictive performance over inference in ecological modelling \cite{breiman2001}. Ensemble methods such as bagging and boosting improve robustness, with RF reducing variance through bootstrapped sampling \cite{breiman2001rf, breiman1996bagging} and XGBoost offering efficient gradient boosting with regularisation \cite{chen2016}. Stacking methods further enhance accuracy by combining diverse learners (e.g., SVM, RF, XGBoost, CNNs) and training a meta-learner on out-of-fold predictions \cite{wolpert1992, breiman1996stacked, friedman2001}. These approaches hold strong potential for automated lobster classification, with applications in fisheries management, aquaculture monitoring, and conservation planning.

\section{Materials and Methodology} \label{sec:proMethod}
\begin{figure}[ht!]
\centering
\includegraphics[width=0.75\textwidth]{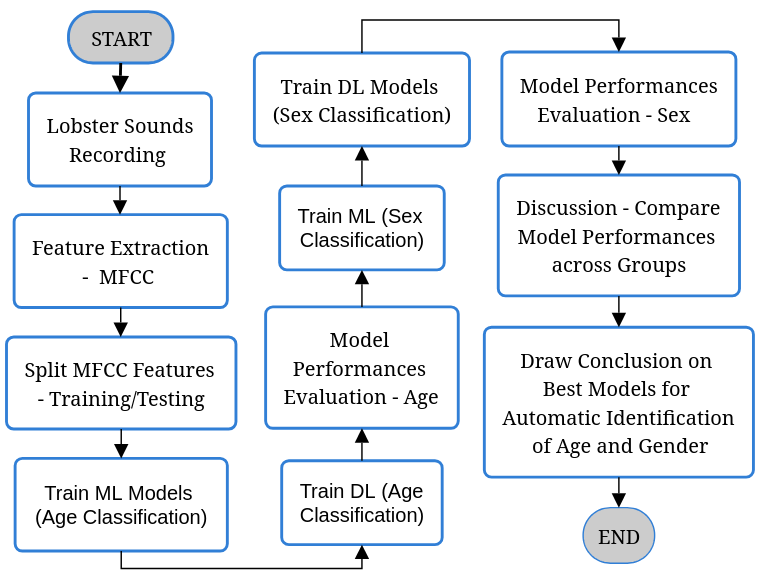}
\caption{Research methodology flowchart depicting the stages undertaken in this study.}
\label{fig:methodologychart}
\end{figure}

This study employed a structured methodology encompassing lobster sound acquisition, preprocessing, feature extraction, dataset preparation, model training with hyperparameter optimisation, performance evaluation, and considerations for edge deployment and reproducibility. The workflow is summarised in Figure~\ref{fig:methodologychart}.

\subsection{Materials Specifications}
Underwater recordings were acquired using an Aquarian Audio H2d hydrophone connected to a Raspberry Pi 3, with audio captured in real time via the PyAudio library. Recordings were subsequently stored and processed on a Lenovo ThinkPad X1 Carbon Gen~13 Aura Edition (PF-2D95WH). Detailed specifications of the Raspberry Pi and hydrophone are provided in Appendix~\ref{sec:appendix}. This hardware configuration balanced portability, real-time acquisition, and computational capability for both data collection and analysis.

\subsection{Study Species and Dataset}
The dataset comprised acoustic recordings of \gls{Hom_euro} (European lobster). In total, 7307~s of recordings were collected from 24 individuals, stratified into four groups: 6 juvenile females, 6 juvenile males, 6 adult males, and 6 adult females. Furthermore, all juvenile were joined (male and female), all adults were joined (male and female); likewise, all male were joined (juvenile and adult) and all female were joined (juvenile and adult). Then, each new class, the long were split into 1 second segments, which were used for the \gls{mfcc} feature extraction for the classification. Table~\ref{tab:lobster_dataset_summary} summarises the dataset by sex and age categories. All recordings were conducted at the Johnshaven Lobster Shop (Scotland) between 16--18 June 2024, with individual sex and age identified visually by experts based on gonopore position and pleopod morphology.

\begin{table}[ht!]
\centering
\caption{Summary of lobster acoustic dataset by sex and age categories.}
\label{tab:lobster_dataset_summary}
\begin{tabular}{|l|c|c|}
\hline
\textbf{Category} & \textbf{Number of Individuals} & \textbf{Recording Duration (s)} \\
\hline
Juvenile Female Lobsters & 6 & 1200 \\
Juvenile Male Lobsters   & 6 & 1200 \\
Adult Male Lobsters      & 6 & 3207 \\
Adult Female Lobsters    & 6 & 1700 \\
\hline
\textbf{Total} & \textbf{24} & \textbf{7307} \\
\hline
\end{tabular}
\end{table}

\subsection{Dataset Recording}\label{sec:dataCollect}
Lobsters were housed in large concrete tanks supplied with continuously circulating seawater, low light, and aeration via pumps. Claws were restrained with rubber bands to prevent aggression \cite{shabani2009spiny}. Concrete tanks were preferred over metal tanks due to reduced reverberation \cite{jezequel2018sound}. Acoustic recordings were acquired at 22.05~kHz, segmented into 1~s files with buffered reads of 6,400 samples. A band-pass filter (50--8,000~Hz) was applied to reduce low-frequency flow noise and high-frequency artefacts.  

\begin{figure}[ht!]
\centering
\includegraphics[width=0.65\textwidth]{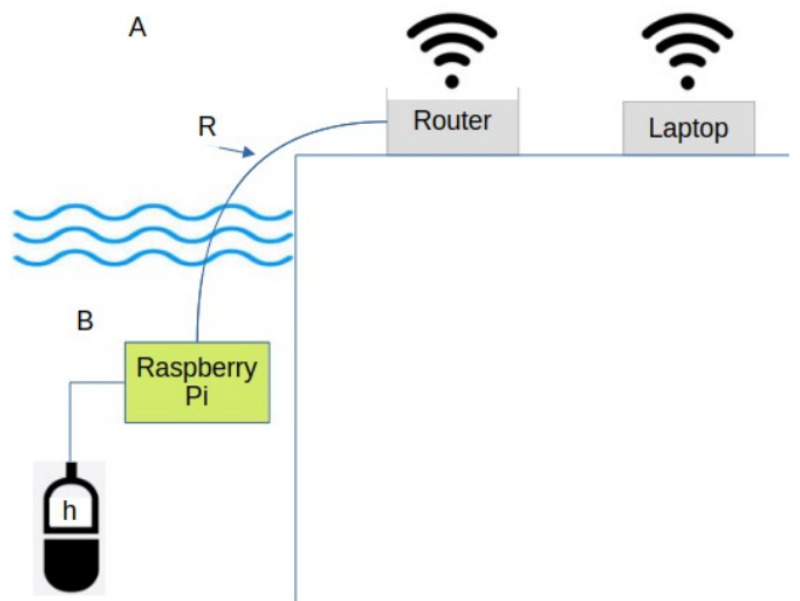}
\caption{Schematic of the acoustic recording system. Area A: airborne environment; Area B: underwater environment where lobster sounds were recorded.}
\label{fig:soundRecordingSystem}
\end{figure}

Sessions ran between 08:00--16:00. Lobsters were not fed during recordings, and environmental conditions (temperature, salinity) were not strictly controlled, ensuring realistic holding conditions. Each lobster was recorded sequentially by age and sex category, then returned to its holding tank. Figure~\ref{fig:soundRecordingSystem} illustrates the recording setup.

\subsection{Data Preprocessing}
Preprocessing ensured consistency and noise reduction prior to feature extraction:
\begin{enumerate}
  \item \textbf{High-pass filtering:} DC offset removed (20--50~Hz).  
  \item \textbf{Band-pass filtering:} 50--8,000~Hz range retained to capture lobster sound energy \cite{jezequel2020acoustic, jezequel2021sound}.  
  \item \textbf{Segmentation:} Fixed 1~s windows.  
  \item \textbf{SNR screening:} Low-energy segments discarded.  
  \item \textbf{Normalisation:} Z-score standardisation applied to all extracted features.  
\end{enumerate}

\subsection{Feature Extraction}
Feature extraction relied exclusively on Mel-Frequency Cepstral Coefficients (MFCCs). For each 1~s segment, 40, 50, and 60 MFCCs were computed using the \texttt{librosa} library and averaged over time, producing a 40-, 50-, and 60-dimensional feature vector. This representation balances robustness to noise with compactness, widely adopted in both speech and bioacoustic classification. Features were standardised prior to model training. No spectrograms or raw waveform inputs were used.

\subsection{Class Balancing and Splitting}
To prevent identity leakage, splitting was performed at the individual level. Data were partitioned into 80\% for training and 20\% for testing, stratified by both sex and age classes. Evaluation metrics were calculated on the held-out 20\% test set. This split ensured sufficient training data while retaining robust generalisation evaluation.

\subsection{Models Evaluated}
Two families of models were implemented for classification:

\paragraph{Classical \gls{ml} models:}
\gls{svm}, \gls{knn}, \gls{nb}, \gls{rf}, \gls{xgboost}, and \gls{mlp}. All were trained on the 40, 50 and 60-dimensional \gls{mfcc} feature vectors, with hyperparameters tuned via cross-validation.

\paragraph{Deep learning model:}
A 1D Convolutional Neural Network (\gls{1d-cnn}) implemented with Keras/TensorFlow. The architecture comprised:
\begin{enumerate}
  \item Conv1D layer (64 filters, kernel size = 3, ReLU).  
  \item MaxPooling1D layer (pool size = 2).  
  \item Flatten layer.  
  \item Dense hidden layer (128 units, ReLU).  
  \item Dense output layer (1 unit, sigmoid) for binary classification.  
\end{enumerate}
Inputs were reshaped MFCC vectors of dimension $(40,1)$. Training employed the Adam optimiser, binary cross-entropy loss, batch size 32, and 10 epochs with early stopping.

\subsection{Performance Metrics}
Performance was assessed using accuracy, precision, recall, F1-score, AUC-ROC, and inference time. Confusion matrices provided error analysis. For rigour, statistical comparisons between models used bootstrap confidence intervals for AUC differences and McNemar tests for paired classification errors, with Benjamini–Hochberg correction applied for multiple comparisons.

\subsection{Edge Deployment Considerations}
Inference feasibility on edge devices was evaluated (e.g., Raspberry Pi, Jetson Nano). Model optimisation techniques such as quantisation, pruning, and knowledge distillation were considered to improve computational efficiency without significant loss of accuracy \cite{sze2017, han2015}. This ensures practical transfer of the pipeline into real-world aquaculture and fisheries monitoring.

\subsection{Reproducibility and Data Availability}
In line with FAIR principles, all raw and processed datasets are publicly available via Mendeley Data \cite{domingos_2025dataset}, with metadata structured according to ISO~19115 standards. Preprocessing and training scripts are openly shared on GitHub \cite{domingos_2025code} to enable full reproducibility.

\section{Experiments and Results}\label{sec:experiment_result}

This study evaluates whether biological attributes (age and sex) can be inferred from lobster bioacoustic recordings using a range of machine learning (ML) and deep learning (DL) methods. Two experiments were carried out:

\begin{enumerate}
  \item \textbf{Experiment 1 — Age-based classification (Adult vs.\ Juvenile).} We compare six classical ML algorithms and eight DL variants (1D-CNN / 1D-DCNN families with varying depth).
  \item \textbf{Experiment 2 — Sex-based classification (Male vs.\ Female).} A comparable set of models and metrics are used to assess discriminability by sex.
\end{enumerate}

For both experiments we report standard discrimination metrics (Accuracy, Precision, Recall, F1-score, AUC-ROC) and per-sample inference time (IT). Inputs are MFCC representations (tested at 40, 50 and 60 coefficients) and, where indicated, features were decorrelated with PCA. All metrics reported are point estimates computed on the held-out test partitions.

\subsection{Experiment 1: Age-based bioacoustics classification}\label{subsec:Exp1}

\begin{figure}[h!]
\centering
\includegraphics[width=1.00\textwidth]{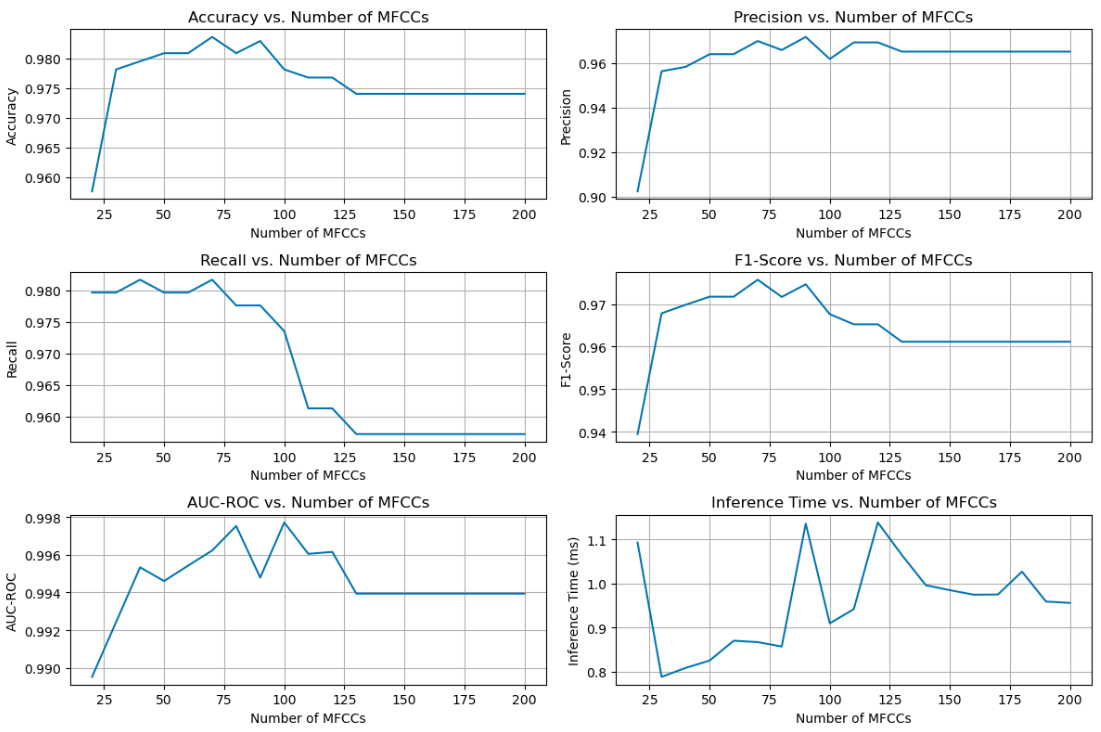}
\caption{Performance metrics for KNN as a function of MFCC dimensionality (Adult vs.\ Juvenile).}
\label{fig:knn_mfcc_graph}
\end{figure}

\begin{figure}[h!]
\centering
\includegraphics[width=1.00\textwidth]{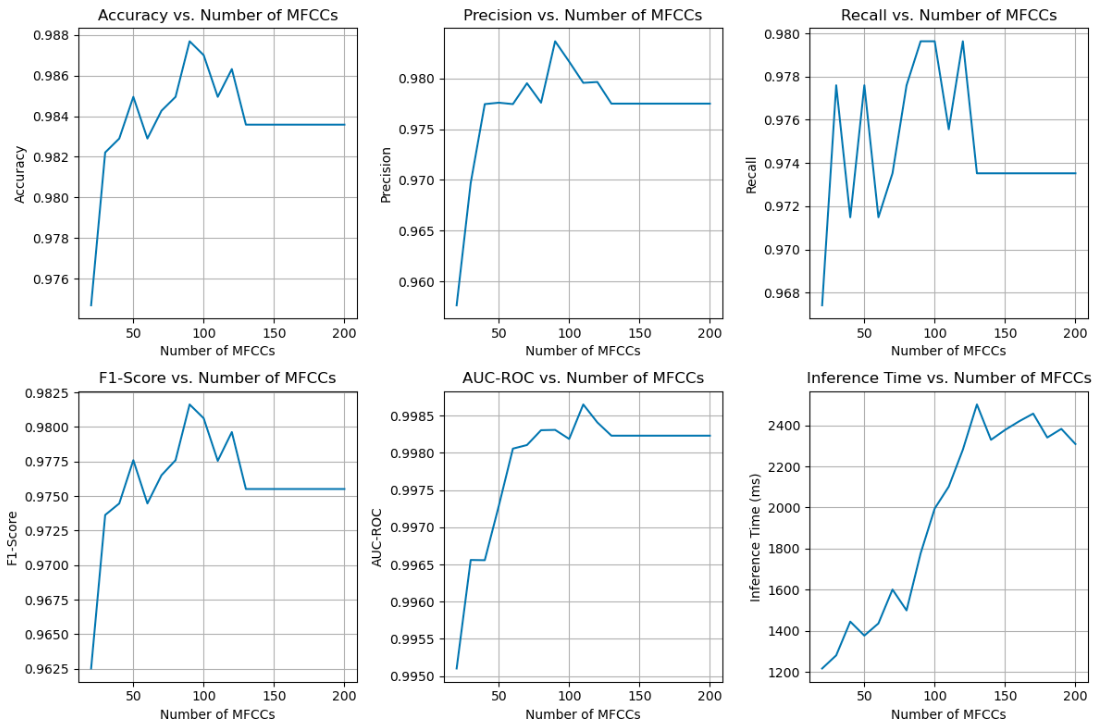}
\caption{Performance metrics for SVM as a function of MFCC dimensionality (Adult vs.\ Juvenile).}
\label{fig:svm_mfcc_graph}
\end{figure}

\begin{figure}[h!]
    \centering
    \begin{minipage}{0.49\textwidth}
        \centering
        \includegraphics[width=\textwidth]{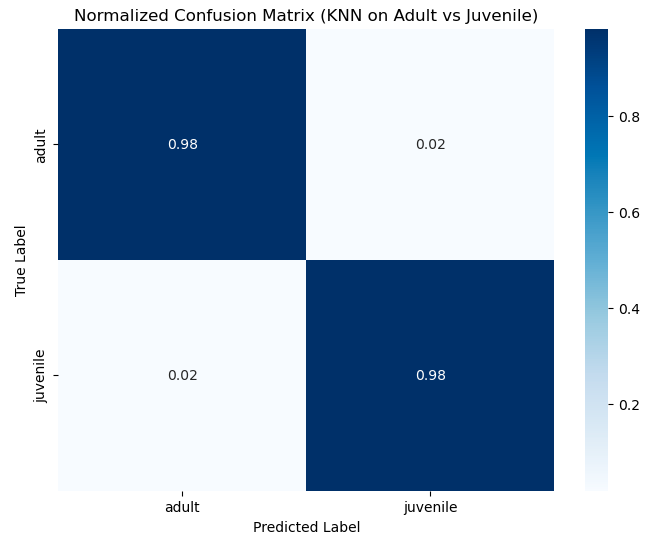}
        \caption{Normalised confusion matrix for KNN (Adult vs.\ Juvenile).}
        \label{fig:knn_confusion_matrix}
    \end{minipage}
    \hfill
    \begin{minipage}{0.49\textwidth}
        \centering
        \includegraphics[width=\textwidth]{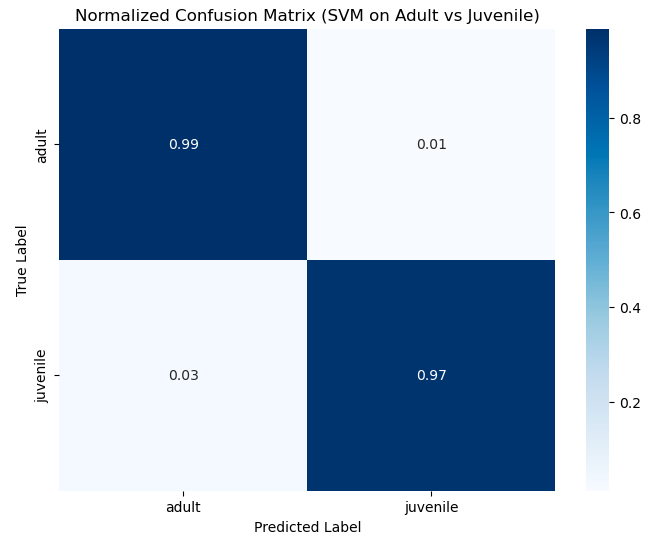}
        \caption{Normalised confusion matrix for SVM (Adult vs.\ Juvenile).}
        \label{fig:svm_confusion_matrix}
    \end{minipage}
\end{figure}

\begin{figure}[h!]
    \centering
    \begin{minipage}{0.49\textwidth}
        \centering
        \includegraphics[width=\textwidth]{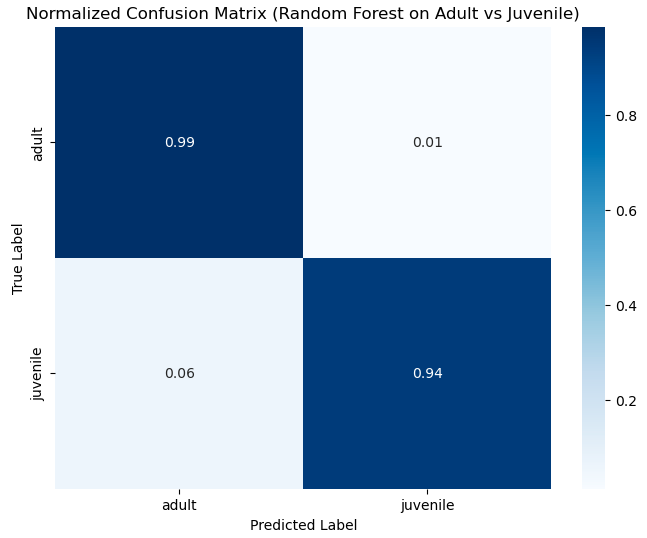}
        \caption{Normalised confusion matrix for RF (Adult vs.\ Juvenile).}
        \label{fig:rf_confusion_matrix}
    \end{minipage}
\end{figure}

\begin{figure}[h!]
    \centering
    \begin{minipage}{0.49\textwidth}
        \centering
        \includegraphics[width=\textwidth]{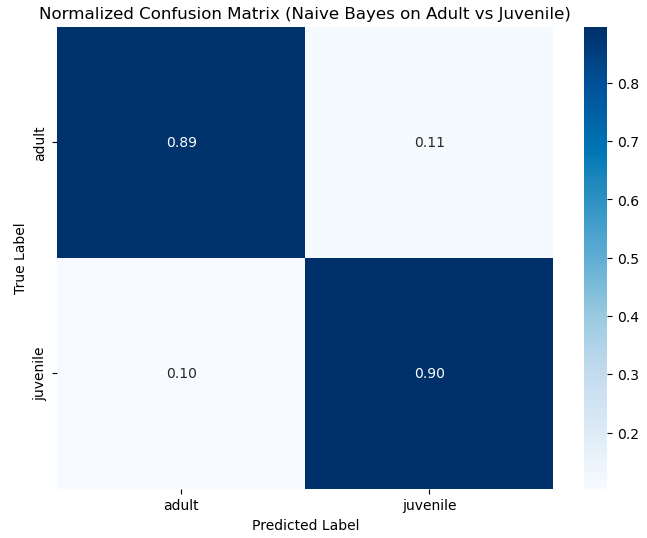}
        \caption{Normalised confusion matrix for NB (Adult vs.\ Juvenile).}
        \label{fig:nb_confusion_matrix}
    \end{minipage}
    \hfill
    \begin{minipage}{0.49\textwidth}
        \centering
        \includegraphics[width=\textwidth]{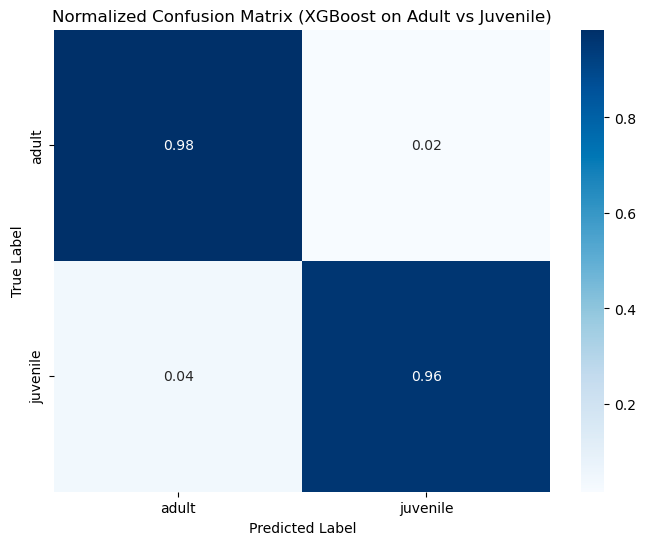}
        \caption{Normalised confusion matrix for XGBoost (Adult vs.\ Juvenile).}
        \label{fig:xgboost_confusion_matrix}
    \end{minipage}
\end{figure}

\begin{figure}[h!]
    \centering
    \begin{minipage}{0.49\textwidth}
        \centering
        \includegraphics[width=\textwidth]{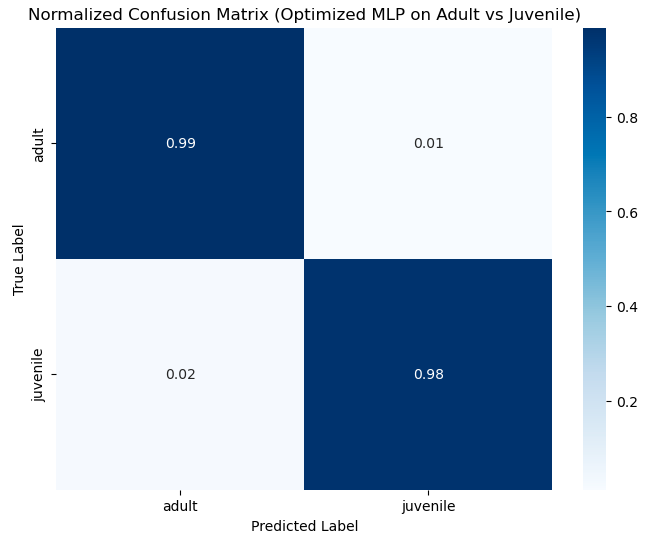}
        \caption{Normalised confusion matrix for MLP (Adult vs.\ Juvenile).}
        \label{fig:mlp_confusion_matrix}
    \end{minipage}
    \hfill
    \begin{minipage}{0.49\textwidth}
        \centering
        \includegraphics[width=\textwidth]{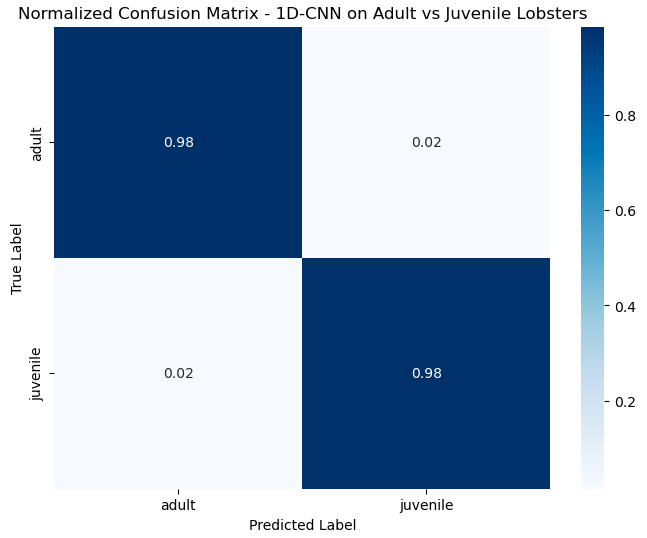}
        \caption{Normalised confusion matrix for 1D-CNN (1 layer) (Adult vs.\ Juvenile).}
        \label{fig:1dccn_1L_confusion_matrix}
    \end{minipage}
\end{figure}

We framed age classification as a binary problem (Adult vs.\ Juvenile) and evaluated six classical \gls{ml} models (\gls{rf}, \gls{xgboost}, \gls{mlp}, \gls{nb}, \gls{svm}, \gls{knn}) and a family of \gls{dl} models (\gls{1d-cnn} and \gls{1d-dcnn} with 1--4 hidden layers). Tables~\ref{tab:ml_adult_juvenile_metrics} and \ref{tab:dl_adult_juvenile_metrics} summarise results across \gls{mfcc} dimensionalities (40, 50, 60). Figures \ref{fig:knn_mfcc_graph} and \ref{fig:svm_mfcc_graph} illustrate metric trends for \gls{knn} and \gls{svm} as \gls{mfcc} number varies; confusion matrices for leading models are presented in Figures~\ref{fig:knn_confusion_matrix}--\ref{fig:mlp_confusion_matrix}.

\subsubsection{Key observations}
\begin{itemize}
  \item \textbf{High discrimination overall.} With the exception of \gls{nb}, nearly all models achieve accuracy $\gtrsim$97\% on the Adult vs.\ Juvenile task.
  \item \textbf{Top performers.} \gls{svm} and \gls{mlp} are among the top performers in several metrics; the highest single accuracy (per the provided point estimates) is \textbf{98.50\%} (\gls{svm}, \gls{mfcc}=50). \gls{mlp} achieves the best \gls{auc-roc} (\textbf{99.86\%}, \gls{mfcc}=60).
  \item \textbf{Inference time trade-offs.} Classical \gls{ml} methods such as \gls{knn} and \gls{nb} yield very low \gls{it} (sub-millisecond to a few milliseconds), whereas \gls{dl} models require seconds per sample — an important operational consideration.
  \item \textbf{Effect of \gls{mfcc} dimensionality.} There is no universal, monotonic improvement with larger \gls{mfcc} counts: model-dependent behaviour is observed (some models improve slightly, others remain unchanged or vary non-monotonically).
\end{itemize}

\begin{table*}[htbp]
  \centering
  \scriptsize
  \caption{Performance metrics of ML models for Adult vs.\ Juvenile classification (MFCC = 40, 50, 60). Best value per column is \textbf{bolded} (for IT lower is better).}
  \label{tab:ml_adult_juvenile_metrics}
  \setlength{\tabcolsep}{10pt}
  \begin{tabularx}{\textwidth}{@{} l c
      S[table-format=2.2] S[table-format=2.2] S[table-format=2.2] S[table-format=2.2] S[table-format=3.2]
      S[table-format=6.2] @{}}
    \toprule
    \textbf{Model} & \textbf{MFCC} & {Accuracy (\%)} & {Precision (\%)} & {Recall (\%)} & {F1 (\%)} & {AUC-ROC (\%)} & {Inference time (ms)} \\
    \midrule

    Random Forest & 40 & 97.20 & 97.27 & 94.30 & 95.76 & 99.50 & 2300.04 \\
                  & 50 & 97.26 & 97.47 & 94.30 & 95.86 & 99.60 & 2511.00 \\
                  & 60 & 97.20 & 97.67 & 93.89 & 95.74 & 99.62 & 2535.04 \\
    \addlinespace

    XGBoost        & 40 & 97.47 & 96.71 & 95.72 & 96.21 & 99.65 & 120.47 \\
                  & 50 & 97.61 & 96.53 & 96.33 & 96.43 & 99.74 & 154.90 \\
                  & 60 & 97.67 & 96.54 & 96.54 & 96.54 & 99.75 & 148.32 \\
    \addlinespace

    MLP            & 40 & 98.43 & \textbf{98.35} & 96.95 & 97.64 & 99.76 & 1271.72 \\
                  & 50 & 97.88 & 96.75 & 96.95 & 96.85 & 99.82 & 1452.11 \\
                  & 60 & 98.43 & 97.37 & 97.96 & 97.66 & \textbf{99.86} & 2015.05 \\
    \addlinespace

    Naive Bayes    & 40 & 89.12 & 80.29 & 89.61 & 84.70 & 97.09 & 2.10 \\
                  & 50 & 91.31 & 82.04 & 94.91 & 88.01 & 97.73 & 2.31 \\
                  & 60 & 90.83 & 80.94 & 95.11 & 87.45 & 97.80 & 2.38 \\
    \addlinespace

    SVM            & 40 & 98.29 & 97.75 & 97.15 & 97.45 & 99.66 & 1381.98 \\
                  & 50 & \textbf{98.50} & 97.76 & \textbf{97.76} & \textbf{97.76} & 99.73 & 1437.78 \\
                  & 60 & 98.29 & 97.75 & 97.15 & 97.45 & 99.81 & 1507.90 \\
    \addlinespace

    KNN            & 40 & 97.95 & 95.83 & \textbf{98.17} & 96.98 & 99.53 & \textbf{0.87} \\
                  & 50 & 98.08 & 96.39 & 97.96 & 97.17 & 99.46 & 0.91 \\
                  & 60 & 98.08 & 96.39 & 97.96 & 97.17 & 99.54 & 1.81 \\

    \bottomrule
  \end{tabularx}

  \vspace{3pt}
  \raggedright\footnotesize\textit{Notes:} Best values per column are bolded.
\end{table*}

\begin{table}[h!]
\centering
\caption{Compact ranking summary (Adult vs Juvenile) — ML models. Lower average rank = better. Ranks computed from point estimates (Accuracy, Precision, Recall, F1, AUC) and Inference Time (IT, lower is better).}
\label{tab:ranking_ml_avj}
\setlength{\tabcolsep}{10pt}
\scriptsize
\begin{tabular}{l c r r r r r r r}
\toprule
Model & MFCC & Acc\_rank & Prec\_rank & Rec\_rank & F1\_rank & AUC\_rank & IT\_rank & AvgRank \\
\midrule
MLP         & 40 & 2 & 1 & 3 & 2 & 1 & 4 & \textbf{2.17} \\
SVM         & 50 & 1 & 2 & 2 & 1 & 3 & 5 & \textbf{2.33} \\
KNN         & 60 & 3 & 5 & 1 & 3 & 5 & 1 & 3.00 \\
XGBoost     & 60 & 4 & 4 & 4 & 4 & 2 & 3 & 3.50 \\
RF          & 50 & 5 & 3 & 6 & 5 & 4 & 6 & 4.83 \\
Naive Bayes & 50 & 6 & 6 & 5 & 6 & 6 & 2 & 5.17 \\
\bottomrule
\end{tabular}
\end{table}

From Table~\ref{tab:ranking_ml_avj} it is conclusive that \gls{mlp} and \gls{svm} tie as best by average rank (\gls{mlp} best by small margin), \gls{knn} and \gls{xgboost} are competitive mid-pack, \gls{rf} lags mainly due to inference time, and \gls{nb} ranks last due to poor discrimination despite fast inference.

\bigskip

The DL family results are shown next. As in the ML table above, best values per column have been highlighted (ties bolded).

\begin{table*}[htbp]
  \centering
  \scriptsize
  \caption{Performance metrics of DL models (1D-CNN and 1D-DCNN families) for Adult vs.\ Juvenile classification. Best value per column is \textbf{bolded} (for IT lower is better).}
  \label{tab:dl_adult_juvenile_metrics}
  \setlength{\tabcolsep}{10pt}
  \begin{tabularx}{\textwidth}{@{} l c
      S[table-format=2.2] S[table-format=2.2] S[table-format=2.2] S[table-format=2.2] S[table-format=3.2]
      S[table-format=6.2] @{}}
    \toprule
    \textbf{Model (layers)} & \textbf{MFCC} & {Accuracy (\%)} & {Precision (\%)} & {Recall (\%)} & {F1 (\%)} & {AUC-ROC (\%)} & {Inference time (ms)} \\
    \midrule

    1D-CNN (1 L)    & 40 & 97.88 & 96.00 & 97.76 & 96.87 & 99.80 & 7537.79 \\
                    & 50 & \textbf{98.15} & 97.15 & 97.35 & \textbf{97.25} & \textbf{99.83} & 16249.91 \\
                    & 60 & 97.67 & 98.11 & 94.91 & 96.48 & 99.79 & 6463.39 \\ 
    \addlinespace

    1D-CNN (2 L)    & 40 & 97.26 & 94.65 & 97.35 & 95.98 & 99.72 & 10580.45 \\
                    & 50 & 97.67 & 96.35 & 96.74 & 96.54 & 99.72 & 27973.32 \\
                    & 60 & 97.40 & 95.03 & 97.35 & 96.18 & 99.72 & 9178.96 \\
    \addlinespace

    1D-CNN (3 L)    & 40 & 97.47 & 95.95 & 96.54 & 96.24 & 99.66 & 13254.90 \\
                    & 50 & 97.40 & 98.09 & 94.09 & 96.05 & 99.59 & 29017.33 \\
                    & 60 & 97.95 & \textbf{98.53} & 95.32 & 96.89 & 99.73 & 12854.35 \\
    \addlinespace

    1D-CNN (4 L)    & 50 & 95.90 & 98.00 & 89.61 & 93.62 & 99.58 & \textbf{3318.51} \\
                    & 60 & 97.67 & 95.07 & \textbf{98.17} & 96.59 & 99.56 & 13406.18 \\
    \addlinespace

    1D-DCNN (1 L)   & 40 & 98.02 & 96.39 & 97.79 & 97.07 & 99.81 & 7478.64 \\
                    & 50 & 98.02 & 96.57 & 97.56 & 97.06 & 99.82 & 8040.94 \\
                    & 60 & 97.74 & 96.54 & 96.74 & 96.64 & \textbf{99.83} & 6583.42 \\
    \addlinespace

    1D-DCNN (2 L)   & 40 & 98.08 & 96.77 & 97.56 & 97.16 & 99.70 & 10615.49 \\
                    & 50 & 97.74 & 95.44 & 97.96 & 96.68 & 99.68 & 12299.25 \\
                    & 60 & 97.47 & 98.30 & 94.09 & 96.15 & 99.58 & 9735.64 \\
    \addlinespace

    1D-DCNN (3 L)   & 40 & 96.24 & 98.23 & 90.43 & 94.17 & 99.54 & 12310.96 \\
                    & 50 & 97.67 & 96.73 & 96.33 & 96.53 & 99.64 & 14671.59 \\
                    & 60 & 96.99 & 93.40 & 97.96 & 95.63 & 99.63 & 12111.80 \\
    \addlinespace

    1D-DCNN (4 L)   & 50 & 95.42 & 91.73 & 94.91 & 93.29 & 99.14 & 15311.84 \\
                    & 60 & 97.13 & 97.06 & 94.30 & 95.66 & 99.69 & 12630.36 \\

    \bottomrule
  \end{tabularx}

  \vspace{3pt}
  \raggedright\footnotesize\textit{Notes:} Best values per column are bolded; ties are bolded for all tied entries. Inference time is per-sample wall-clock time measured on the tuning hardware.
\end{table*}

\begin{table}[h!]
\centering
\caption{Compact ranking summary (Adult vs Juvenile) — DL models (1D-CNN / 1D-DCNN families). Lower average rank = better.}
\label{tab:ranking_dl_avj}
\setlength{\tabcolsep}{10pt}
\scriptsize
\begin{tabular}{l c r r r r r r r}
\toprule
Model (layers) & MFCC & Acc\_rank & Prec\_rank & Rec\_rank & F1\_rank & AUC\_rank & IT\_rank & AvgRank \\
\midrule
1D-CNN (1 L)   & 50 & 1 & 2 & 4 & 1 & 1 & 7 & \textbf{2.67} \\
1D-DCNN (1 L)  & 40 & 3 & 6 & 2 & 3 & 2 & 1 & \textbf{2.83} \\
1D-DCNN (2 L)  & 40 & 2 & 4 & 3 & 2 & 5 & 2 & 3.00 \\
1D-CNN (3 L)   & 60 & 4 & 1 & 7 & 4 & 3 & 4 & 3.83 \\
1D-CNN (4 L)   & 60 & 6 & 8 & 1 & 5 & 8 & 5 & 5.50 \\
1D-CNN (2 L)   & 50 & 6 & 7 & 5 & 6 & 4 & 8 & 6.00 \\
1D-DCNN (4 L)  & 60 & 8 & 3 & 8 & 8 & 6 & 3 & 6.00 \\
1D-DCNN (3 L)  & 50 & 6 & 5 & 6 & 7 & 7 & 6 & 6.17 \\
\bottomrule
\end{tabular}
\end{table}

Based on the results in Table~\ref{tab:ranking_dl_avj}, \gls{1d-cnn} (1 layer) and \gls{1d-dcnn} (1–2 layers) yield the best average rank: good discrimination with comparatively lower inference costs among \gls{dl}s. Very deep variants (3–4 layers) rarely yield substantial gains but increase inference latency.

\subsection{Experiment 1: Comparative analysis}\label{subsec:analysis1}

Overall, classical ML and shallow DL models both achieve strong performance on the adult vs.\ juvenile task. Key takeaways:

\begin{itemize}
  \item \textbf{SVM and MLP are among the most consistent classifiers} with top-tier accuracy and F1; SVM reaches the top accuracy point while MLP yields the top AUC.
  \item \textbf{KNN has excellent recall and the fastest inference time} (suitable for low-latency systems), while Naive Bayes is computationally cheap but substantially weaker in accuracy.
  \item \textbf{DL models can match or slightly exceed ML performance} (notably some 1D-DCNN variants), but generally at a significantly higher inference cost.
  \item \textbf{MFCC dimensionality matters but is model dependent.} We cannot recommend a single MFCC count for all models; nested CV and broader sweeps (e.g., 20–80) are advised.
\end{itemize}

\bigskip

\subsection{Experiment 2: Sex-based bioacoustics classification}

\begin{figure}[h!]
\centering
\includegraphics[width=\textwidth]{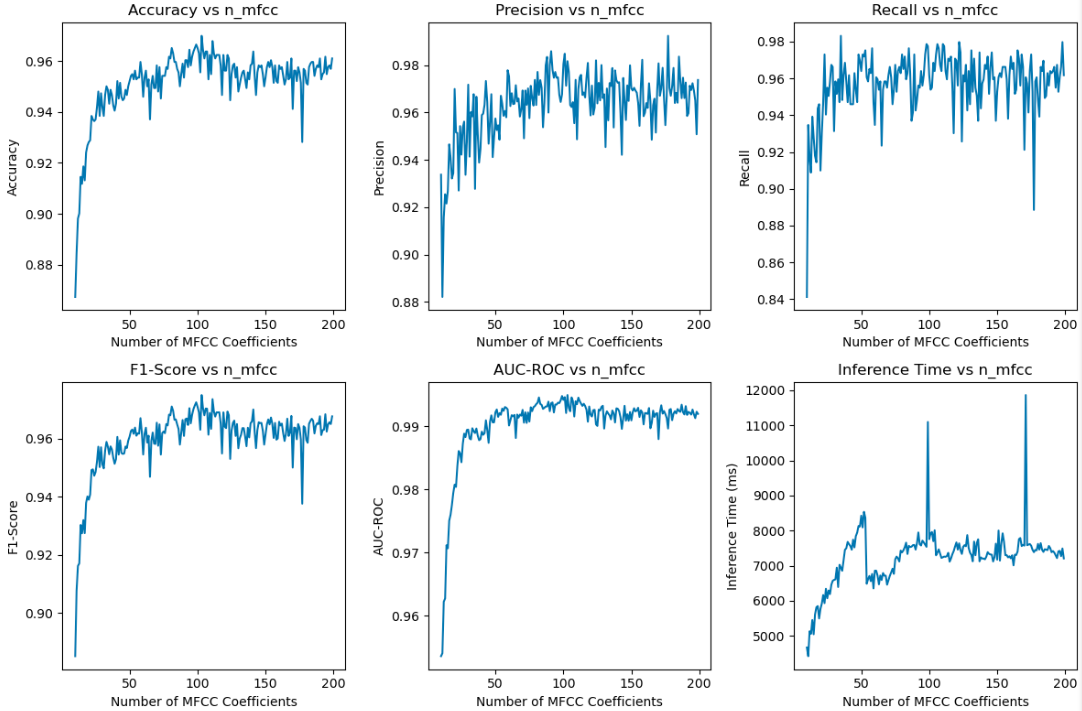}
\caption{Performance of 1D-CNN (1 hidden layer) as a function of MFCC dimensionality for male vs. female classification. Points show mean performance per metric (accuracy, precision, recall, F1, AUC-ROC, inference time) computed across cross-validation folds.}
\label{fig:metric_graph_1dccn1l}
\end{figure}

We evaluated the capacity of machine-learning (ML) and deep-learning (DL) models to discriminate sex (male vs. female) from lobster bioacoustic recordings. Models tested include Random Forest (RF), XGBoost, multilayer perceptron (MLP), Naive Bayes (NB), support vector machine (SVM), K-nearest neighbours (KNN) and a family of 1D convolutional networks (1D-CNN and 1D-DCNN) with varying depths. All inputs were derived from MFCC representations (40, 50 and 60 coefficients) and preprocessed with PCA where indicated. Performance is summarised in Table~\ref{tab:ml_male_female_metrics} (ML classifiers) and Table~\ref{tab:dl_male_female_metrics} (DL classifiers). Figure~\ref{fig:metric_graph_1dccn1l} illustrates metric trends for the 1D-CNN (1 hidden layer) across MFCC dimensions.

\begin{table*}[htbp]
  \centering
  \scriptsize
  \caption{Performance metrics of ML models for Male vs.\ Female classification (MFCC = 40, 50, 60). Best value per column is \textbf{bolded}.}
  \label{tab:ml_male_female_metrics}
  \setlength{\tabcolsep}{11pt}
  \begin{tabularx}{\textwidth}{@{} l c
      S[table-format=2.2] S[table-format=2.2] S[table-format=2.2] S[table-format=2.2] S[table-format=3.2]
      S[table-format=6.4] @{}}
    \toprule
    \textbf{Model} & \textbf{MFCC} & {Accuracy (\%)} & {Precision (\%)} & {Recall (\%)} & {F1 (\%)} & {AUC-ROC (\%)} & {Inference time (ms)} \\
    \midrule

    Random Forest & 40 & 93.23 & 94.58 & 94.26 & 94.42 & 98.08 & 2559.0391 \\
                  & 50 & 92.54 & 93.23 & 94.59 & 93.91 & 98.16 & 2986.6513 \\
                  & 60 & 92.54 & 93.33 & 94.48 & 93.90 & 98.18 & 2968.7502 \\
    \addlinespace

    XGBoost        & 40 & 94.12 & 94.75 & 95.61 & 95.18 & 98.73 & 334.6043 \\
                  & 50 & 95.21 & 96.16 & 95.95 & 96.05 & 98.81 & 231.1075 \\
                  & 60 & 95.28 & 96.17 & 96.06 & 96.11 & 98.97 & 221.2673 \\
    \addlinespace

    MLP            & 40 & 95.35 & 96.80 & 95.50 & 96.15 & 99.03 & 3324.9233 \\
                  & 50 & 95.14 & 96.58 & 95.38 & 95.98 & \textbf{99.19} & 2794.5994 \\
                  & 60 & 95.90 & 96.83 & 96.40 & 96.61 & 99.16 & 2260.6717 \\
    \addlinespace

    Naive Bayes    & 40 & 81.60 & 87.52 & 81.31 & 84.30 & 91.04 & 2.3317 \\
                  & 50 & 82.56 & 88.36 & 82.09 & 85.11 & 90.85 & 2.0922 \\
                  & 60 & 82.01 & 88.82 & 80.52 & 84.47 & 89.83 & 2.3787 \\
    \addlinespace

    SVM            & 40 & 95.42 & 96.07 & 96.40 & 96.23 & 98.86 & 2586.4148 \\
                  & 50 & \textbf{96.17} & 96.95 & \textbf{96.73} & \textbf{96.84} & 99.03 & 2609.8358 \\
                  & 60 & 95.96 & 96.94 & 96.40 & 96.67 & 99.11 & 2811.5256 \\
    \addlinespace

    KNN            & 40 & 94.87 & 95.93 & 95.61 & 95.77 & 98.85 & 0.8528 \\
                  & 50 & 95.49 & 97.03 & 95.50 & 96.25 & 98.97 & \textbf{0.6002} \\
                  & 60 & 95.62 & \textbf{97.14} & 95.61 & 96.37 & 99.00 & 0.6658 \\

    \bottomrule
  \end{tabularx}

  \vspace{3pt}
  \raggedright\footnotesize\textit{Notes:} Best per-column values are bolded (for inference time lower is better).
\end{table*}

\begin{table}[h!]
\centering
\caption{Compact ranking summary (Male vs Female) --- ML models. Lower average rank = better. Cells shaded indicate rank $\leq$ 2 (top-2).}
\label{tab:ranking_ml_mf}
\setlength{\tabcolsep}{11pt}
\scriptsize
\begin{tabular}{l c r r r r r r r}
\toprule
Model & MFCC & Acc\_rank & Prec\_rank & Rec\_rank & F1\_rank & AUC\_rank & IT\_rank & AvgRank \\
\midrule
KNN            & 60 & 3 & \cellcolor{gray!25}1 & 4 & 3 & 3 & \cellcolor{gray!25}1 & 2.50 \\
MLP            & 60 & \cellcolor{gray!25}2 & 3 & \cellcolor{gray!25}2 & \cellcolor{gray!25}2 & \cellcolor{gray!25}1 & 4 & \textbf{2.33} \\
Naive Bayes    & 50 & 6 & 6 & 6 & 6 & 6 & \cellcolor{gray!25}2 & 5.33 \\
Random Forest  & 40 & 5 & 5 & 5 & 5 & 5 & 5 & 5.00 \\
SVM            & 50 & \cellcolor{gray!25}1 & \cellcolor{gray!25}2 & \cellcolor{gray!25}1 & \cellcolor{gray!25}1 & \cellcolor{gray!25}2 & 6 & \textbf{2.17} \\
XGBoost        & 60 & 4 & 4 & 3 & 4 & 4 & 3 & 3.67 \\
\bottomrule
\end{tabular}
\end{table}

The ranking summary in Table~\ref{tab:ranking_ml_mf} indicates that for the classification of Male vs Female \gls{svm} and \gls{mlp} are the strongest performers by average rank (\gls{svm} marginally best on accuracy and multiple metrics; MLP achieves top AUC), while \gls{knn} is notable for best precision and best (lowest) inference time. \gls{xgboost} and \gls{rf} occupy the mid-range; \gls{nb} ranks lowest due to substantially weaker discrimination despite very low IT.

\begin{table*}[htbp]
  \centering
  \scriptsize
  \caption{Performance metrics of DL models (1D-CNN and 1D-DCNN families) for Male vs.\ Female classification. Best value per column is \textbf{bolded}.}
  \label{tab:dl_male_female_metrics}
  \setlength{\tabcolsep}{10pt}
  \begin{tabularx}{\textwidth}{@{} l c
      S[table-format=2.2] S[table-format=2.2] S[table-format=2.2] S[table-format=2.2] S[table-format=3.2]
      S[table-format=6.4] @{}}
    \toprule
    \textbf{Model (layers)} & \textbf{MFCC} & {Accuracy (\%)} & {Precision (\%)} & {Recall (\%)} & {F1 (\%)} & {AUC-ROC (\%)} & {Inference time (ms)} \\
    \midrule

    1D-CNN (1 L)    & 40 & 94.60 & 94.50 & 96.73 & 95.60 & 98.75 & 8253.8011 \\
                    & 50 & 94.75 & 95.40 & 96.03 & 95.69 & 99.04 & 8069.7830 \\
                    & 60 & \textbf{95.35} & 95.61 & 96.82 & \textbf{96.20} & 99.13 & \textbf{6433.0564} \\ 
    \addlinespace

    1D-CNN (2 L)    & 40 & 94.25 & 95.17 & 95.38 & 95.28 & 98.67 & 10445.2045 \\
                    & 50 & 95.04 & 96.09 & \textbf{97.78} & 95.92 & 98.95 & 12180.0492 \\
                    & 60 & 94.94 & 95.83 & 95.86 & 95.83 & 99.00 & 8758.4734 \\
    \addlinespace

    1D-CNN (3 L)    & 40 & 92.13 & 90.64 & 97.07 & 93.75 & 98.10 & 13290.2761 \\
                    & 50 & 93.61 & 94.98 & 94.51 & 94.72 & 98.64 & 15455.8352 \\
                    & 60 & 94.63 & 96.05 & 95.08 & 95.56 & 98.78 & 12078.9611 \\
    \addlinespace

    1D-CNN (4 L)    & 50 & 93.66 & 93.45 & 96.31 & 94.86 & 98.49 & 15676.7719 \\
                    & 60 & 93.93 & 94.69 & 95.39 & 95.02 & 98.62 & 13249.9224 \\
    \addlinespace

    1D-DCNN (1 L)   & 40 & 94.19 & 94.66 & 95.83 & 95.24 & 98.79 & 7332.1474 \\
                    & 50 & \textbf{95.35} & 95.44 & 96.99 & \textbf{96.20} & 99.19 & 7940.4736 \\
                    & 60 & 95.11 & 95.70 & 96.37 & 96.00 & \textbf{99.21} & 6515.3026 \\
    \addlinespace

    1D-DCNN (2 L)   & 40 & 93.91 & 96.26 & 93.58 & 94.92 & 98.47 & 10611.2289 \\
                    & 50 & 94.87 & 95.75 & 95.83 & 95.76 & 98.97 & 12129.8255 \\
                    & 60 & 95.03 & 96.01 & 95.83 & 95.85 & 99.05 & 8682.1848 \\
    \addlinespace

    1D-DCNN (3 L)   & 40 & 92.95 & 93.47 & 95.05 & 94.25 & 98.27 & 12144.5677 \\
                    & 50 & 93.98 & 94.45 & 95.72 & 95.08 & 98.58 & 14186.0330 \\
                    & 60 & 94.27 & 95.35 & 95.22 & 95.28 & 98.63 & 11781.5477 \\
    \addlinespace

    1D-DCNN (4 L)   & 50 & 93.23 & \textbf{98.80} & 93.92 & 94.40 & 98.46 & 15143.1859 \\
                    & 60 & 92.89 & 96.34 & 91.78 & 94.00 & 98.37 & 12650.9302 \\

    \bottomrule
  \end{tabularx}

  \vspace{3pt}
  \raggedright\footnotesize\textit{Notes:} Best per-column values are bolded (for inference time lower is better). Ties are bolded for all tied entries.
\end{table*}

\begin{table}[h!]
\centering
\caption{Compact ranking summary (Male vs Female) --- DL models (1D-CNN / 1D-DCNN). Lower average rank = better. Cells shaded indicate rank $\leq$ 2 (top-2).}
\label{tab:ranking_dl_mf}
\setlength{\tabcolsep}{11pt}
\scriptsize
\begin{tabular}{l c r r r r r r r}
\toprule
Model (layers) & MFCC & Acc\_rank & Prec\_rank & Rec\_rank & F1\_rank & AUC\_rank & IT\_rank & AvgRank \\
\midrule
1D-CNN (1 L)    & 60 & \cellcolor{gray!25}1 & 5 & 3 & \cellcolor{gray!25}1 & \cellcolor{gray!25}2 & \cellcolor{gray!25}1 & \textbf{2.17} \\
1D-CNN (2 L)    & 50 & 3 & \cellcolor{gray!25}2 & \cellcolor{gray!25}1 & 3 & 4 & 6 & 3.17 \\
1D-CNN (3 L)    & 60 & 5 & 3 & 7 & 5 & 5 & 5 & 5.00 \\
1D-CNN (4 L)    & 60 & 7 & 8 & 5 & 7 & 7 & 7 & 6.83 \\
1D-DCNN (1 L)   & 50 & \cellcolor{gray!25}1 & 6 & \cellcolor{gray!25}2 & \cellcolor{gray!25}1 & \cellcolor{gray!25}1 & \cellcolor{gray!25}2 & \textbf{2.17} \\
1D-DCNN (2 L)   & 60 & 4 & 4 & 4 & 4 & 3 & 3 & 3.67 \\
1D-DCNN (3 L)   & 60 & 6 & 7 & 6 & 6 & 6 & 4 & 5.83 \\
1D-DCNN (4 L)   & 50 & 8 & \cellcolor{gray!25}1 & 8 & 8 & 8 & 8 & 6.83 \\
\bottomrule
\end{tabular}
\end{table}

The results in Table~\ref{tab:ranking_dl_mf} summarises that among \gls{dl}s, shallow architectures (\gls{1d-cnn} 1-layer and \gls{1d-dcnn} 1-layer) obtain the best average ranks, balancing strong discrimination and comparatively lower inference costs. Some deeper architectures have high precision but suffer from larger inference time; deeper does not uniformly improve the overall rank.

\subsubsection{Summary of main findings}
Across the classical \gls{ml} classifiers, \gls{svm} achieved the highest point accuracy (96.17\%, MFCC=50) and high values on other discrimination metrics (precision, recall, F1) (see Table~\ref{tab:ml_male_female_metrics}). \gls{xgboost} and \gls{mlp} are close competitors: \gls{xgboost} attains accuracies above 95\% for \gls{mfcc}=50 and \gls{mfcc}=60, while the \gls{mlp} likewise attains its best accuracy at \gls{mfcc}=60 (95.90\%). \gls{nb} performs substantially worse than the other ML methods (\(\approx\)82\%–83\% accuracy), illustrating that strong parametric assumptions are less appropriate for this task.

For the \gls{dl} family, shallow architectures (1–2 hidden layers) generally match or slightly outperform the best ML models on overall accuracy and \gls{auc-roc}, with the \gls{1d-cnn} and \gls{1d-dcnn} achieving peak accuracies of 95.35\% under different \gls{mfcc} settings (Table~\ref{tab:dl_male_female_metrics}). \gls{auc-roc} values are uniformly high across DL models (>98\%), indicating strong separability between male and female classes.

\gls{it} differs substantially between paradigms: \gls{ml} classifiers (except \gls{rf} which uses many trees) generally yield low inference latency relative to \gls{dl} models; \gls{dl} \gls{it}s are several seconds per sample (Tables~\ref{tab:ml_male_female_metrics},~\ref{tab:dl_male_female_metrics}), which may constrain real-time deployments.

\subsubsection{Interpretation and practical implications}
The results indicate three practical points:

\begin{enumerate}
  \item \textbf{Top accuracy vs. operational cost.} SVM and XGBoost produce the best discrimination with modest inference cost (SVM is competitive albeit with higher IT than lightweight ML methods). Deep models yield comparable discrimination but incur substantially higher inference times; this trade-off should weigh in deployment decisions where latency matters.
  \item \textbf{Model-dependent sensitivity to input dimensionality.} Increasing MFCC count does not universally improve performance. For some models (XGBoost, MLP) modest gains occur when moving from 40 to 50/60 MFCCs; for RF and some CNN variants the effect is neutral or inconsistent. Thus, optimal MFCC dimensionality appears model-dependent and should be identified per algorithm (preferably by nested CV).
  \item \textbf{Shallow architectures can suffice.} For both 1D-CNN and 1D-DCNN families the best performance is often achieved with one or two hidden layers; deeper architectures yield diminishing returns and substantially larger inference times, suggesting overparameterisation for this dataset.
\end{enumerate}

\subsubsection{On inference time and deployment}
DL models exhibit inference times in the multi-second range per sample (Tables~\ref{tab:dl_male_female_metrics}), whereas some ML models (e.g. KNN, NB) have very low latency. When deployment latency is critical (e.g., in-situ monitoring, edge devices), we recommend profiling models under representative hardware and considering:
\begin{itemize}
  \item Model compression / quantisation for CNNs.
  \item Pruning or smaller ensembles for tree/boosting methods.
  \item Lightweight feature sets selected via feature-importance (SHAP) or sequential feature selection.
\end{itemize}

\subsubsection{Limitations}
\begin{itemize}
  \item Only three MFCC dimensionalities were evaluated (40, 50, 60). A broader sweep (e.g., 20–80 in 10-step increments) would better characterise the relationship between feature dimensionality and performance.
  \item Reported metrics appear as point estimates in Tables~\ref{tab:ml_male_female_metrics} and~\ref{tab:dl_male_female_metrics}.
  \item Inference time measurements can be confounded by background processes and hardware variability; report hardware specs (CPU/GPU, RAM) and repeat timing with controlled conditions (fixed CPU affinity, disabled turbo boost) or present median times over multiple runs.
\end{itemize}

\subsubsection{Summary and recommendations}
SVM and XGBoost are the best performing classical approaches on this dataset; shallow DL architectures are competitive but more computationally expensive at inference. We therefore recommend:
\begin{itemize}
  \item Reporting uncertainty (CIs) and formal statistical comparisons (McNemar / bootstrap / DeLong) in the final manuscript.
  \item Prioritising models for deployment based on the accuracy–latency trade-off, and applying calibration corrections when probabilistic outputs are needed.
  \item Performing additional MFCC dimensionality sweeps and nested CV to robustly identify model-dependent optima.
\end{itemize}

\subsection{Stacking ensemble: proposed pipeline and rationale}

Ensemble learning synthesises multiple base learners to produce a single, often more accurate and robust predictor than any individual model \cite{sill2009feature,zhou2012ensemble,breiman1996stacked,wolpert1992,dietterich2000ensemble,friedman2001}. Following this principle, we propose a stacking ensemble that explicitly leverages complementary model inductive biases to improve discrimination and calibration on bioacoustic classification tasks.

\paragraph{Design rationale.}
We assemble a diverse set of base learners—a Random Forest (\gls{rf}), an XGBoost classifier (\gls{xgboost}), a support vector machine (\gls{svm}) and a 1D convolutional neural network (\gls{1d-cnn})—because (i) tree ensembles (RF, XGBoost) provide complementary variance- and bias-reduction mechanisms \cite{breiman2001rf,chen2016}, (ii) SVMs often perform well on moderate-dimensional engineered feature spaces, and (iii) 1D-CNNs can extract local temporal/spectral patterns directly from short waveform or spectrogram segments. Combining these heterogeneous learners increases the likelihood that different aspects of the signal are captured, which stacking can exploit \cite{wolpert1992,breiman1996stacked}.

\paragraph{Data streams and feature engineering.}
The pipeline maintains two parallel feature streams: (1) engineered tabular features (spectral summaries, statistical descriptors and other domain-specific features), used by RF, XGBoost and SVM; and (2) raw or minimally preprocessed short waveform / spectrogram segments used to train the 1D-CNN. Prior to modelling, tabular features are standardised and, where appropriate, decorrelated with PCA; PCA is only applied within training folds to avoid leakage.

\paragraph{Training protocol and leakage prevention.}
To avoid target leakage and to produce out-of-sample meta-training data, we use stratified $K$-fold cross-validation on the training set (here $K=5$ unless otherwise stated) to generate out-of-fold (OOF) predicted probabilities for each base learner. For each fold:
\begin{enumerate}
  \item Train each base learner on the $K-1$ folds (using the same preprocessing pipeline fit only on those folds).
  \item Predict class probabilities on the held-out fold and store these OOF probabilities.
\end{enumerate}
After completing $K$ folds we obtain an OOF prediction matrix of shape $(N_{\text{train}}, B \times C)$ where $B$ is the number of base learners and $C$ the number of classes. These OOF features are then used to train a low-capacity meta-learner (logistic regression with $L_2$ regularisation by default) that maps the base learners' probability outputs to final predictions. Finally, for deployment each base learner is retrained on the full training set (with preprocessing pipelines refit on full training data) and the meta-learner is used to combine their outputs at test time. This procedure prevents information flow from test folds into the meta-learner during training and is standard in stacking implementations \cite{wolpert1992,breiman1996stacked}.
 
\paragraph{Meta-learner and calibration.}
We recommend a low-capacity, regularised meta-learner (logistic regression or a small gradient-boosting machine) to avoid overfitting the OOF matrix. Because stacked outputs can be miscalibrated, we emphasise post-hoc calibration diagnostics and correction (e.g., reliability diagrams, Brier score, Platt scaling or isotonic regression) as part of the validation pipeline \cite{breiman2001rf}. Where probabilistic interpretation is required, calibration should be reported alongside discrimination metrics.

\paragraph{Practical considerations.}
We recommend:
\begin{itemize}
  \item Hyperparameter optimisation for base learners inside the inner cross-validation loop (or via nested CV) to avoid optimistic bias.
  \item Parallelised training of base learners and careful resource reporting (CPU/GPU, wall-clock time) to document computational cost.
  \item Storing OOF predictions, fitted preprocessing objects, random seeds, and software environment (package versions or containers) to ensure reproducibility according to FAIR principles.
  \item Interpreting ensemble behaviour using feature-importance measures (e.g. SHAP for tree-based learners), and inspecting meta-learner coefficients to understand how base learners are weighted.
\end{itemize}

\begin{figure}[htbp]
  \centering
  \includegraphics[width=0.9\textwidth]{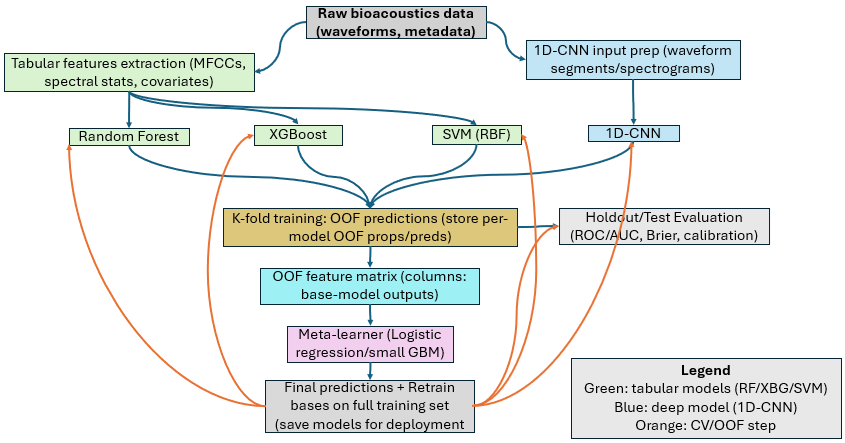}
  \caption{Schematic of the proposed stacking ensemble pipeline. Two parallel feature streams (engineered tabular features and 1D-CNN-ready waveform/spectrogram segments) are processed by diverse base learners. Out-of-fold (OOF) probability outputs from each base learner form the meta-feature matrix used to train a low-capacity meta-learner; final predictions are produced by retraining base learners on the full training set and combining their outputs through the fitted meta-learner.}
  \label{fig:stacking_placeholder}
\end{figure}

\paragraph{Implementation note.}
A schematic of the proposed stacking pipeline is shown in Figure~\ref{fig:stacking_placeholder}. The implementation (data preprocessing pipelines, model definitions, and OOF-generation scripts) is provided in the code repository cited in the manuscript \cite{domingos2025stacking}. We also provide recommended default meta-learner settings (logistic regression with $L_2$ penalty, regularisation strength selected by CV) and suggest that authors include a brief ablation table in supplementary materials comparing averaging, majority-vote, and stacked meta-learner performance.

\section{Model architectures and parametrisation}

A clear, concise description of model architectures and their parametrisation is essential for scientific reproducibility and interpretability. For each method studied we document the architecture (where applicable), the search space explored during hyperparameter tuning, and the final selected settings. We also report computational costs for hyperparameter optimisation to provide practical guidance for reproducibility and scale-up. The following subsections summarise the optimisation protocol and the best configurations found for each algorithm applied to \gls{mfcc}-derived features (preprocessed with principal component analysis where indicated).

\subsection{KNN parameter optimisation}

We optimised the $k$-nearest neighbours (KNN) classifier via grid search across plausible neighbourhood sizes and distance metrics. Input features were derived from \gls{mfcc} coefficients; principal component analysis (\gls{pca}) was applied primarily for decorrelation and to stabilise optimisation — the retained component count and cumulative total explained variance (TEV) are reported below. 

Table~\ref{tab:knn_arc} summarises the results for three initial \gls{mfcc} configurations. For each case we report TEV after PCA, the neighbour-search algorithm, the selected number of neighbours ($k$), the Minkowski exponent ($p$), the weight function used for voting, the mean cross-validated test accuracy (expressed as \%), and the wall-clock time required for grid search. The results indicate a modest improvement when increasing the input MFCCs from 40 to 50 (97.72\% → 97.79\%), with negligible additional computational cost; further increasing to 60 MFCCs did not yield further gains. These findings support the use of PCA for compact, decorrelated representations and suggest that modest increases in MFCC dimensionality can improve KNN performance for this dataset.

\begin{table*}[htbp]
  \centering
  \scriptsize
  \caption{KNN grid-search and PCA settings. PCA was applied to decorrelate MFCC features; TEV is the cumulative total explained variance by the retained components.}
  \label{tab:knn_arc}
  \setlength{\tabcolsep}{6pt}
  \begin{tabularx}{\textwidth}{@{} 
      l  
      c  
      p{1.6cm} 
      p{1.6cm} 
      S[table-format=1.2] 
      c  
      c  
      c  
      c  
      S[table-format=2.2] 
      S[table-format=3.2] 
      @{}}
    \toprule
    \textbf{Model} & \textbf{\# MFCC} & \textbf{Original shape} & \textbf{Shape after PCA} &
    \textbf{TEV} & \textbf{Algorithm} & \textbf{\# Neighbors} & \textbf{$p$} & \textbf{Weight} & \textbf{Best CV Acc. (\%)} & \textbf{Grid time (s)} \\
    \midrule

    KNN & 40 & (7307, 40) & (7307, 40) & 1.00 & auto & 5 & 1 & uniform & 97.72 & 10.93 \\
    KNN & 50 & (7307, 50) & (7307, 40) & 0.98 & auto & 7 & 2 & uniform & 97.79 & 12.10 \\
    KNN & 60 & (7307, 60) & (7307, 40) & 0.96 & auto & 9 & 1 & uniform & 97.72 & 12.21 \\

    \bottomrule
  \end{tabularx}

  \vspace{3pt}
  \raggedright\footnotesize\textit{Notes:} `Algorithm' is the neighbour-search method passed to \texttt{scikit-learn}'s \texttt{KNeighborsClassifier}; `$p$' is the Minkowski exponent (1 = Manhattan, 2 = Euclidean). Best CV accuracy is the mean over cross-validation folds (reported as percent). Grid time is wall-clock tuning time on the reported machine.
\end{table*}

\subsection{SVM parameter optimisation}
We performed a grid search to identify optimal support vector machine (SVM) hyperparameters for classification using \gls{mfcc}-derived features preprocessed with principal component analysis (\gls{pca}). PCA was applied to decorrelate features and reduced each input set to 30 components while preserving a high cumulative total explained variance (TEV). Table~\ref{tab:svm_arc} summarises the best validation configurations for each initial number of \gls{mfcc} coefficients. For each case we report TEV after PCA and the hyperparameters that yielded the best cross-validated performance (regularisation parameter $C$, kernel scale `gamma`, and kernel type). Reported grid-search wall-clock times provide an empirical indication of tuning cost. These results indicate the SVM is robust across input dimensionalities, with only modest adjustments to regularisation and kernel scaling required.

\begin{table*}[htbp]
  \centering
  \scriptsize
  \caption{SVM grid-search and PCA settings. PCA reduced feature dimensionality to 30 components while retaining high TEV.}
  \label{tab:svm_arc}
  \setlength{\tabcolsep}{6pt}
  \begin{tabularx}{\textwidth}{@{}
      l 
      c 
      p{1.8cm} 
      p{1.8cm} 
      S[table-format=1.2] 
      p{1.6cm} 
      p{1.6cm} 
      c 
      S[table-format=5.2] 
      @{}}
    \toprule
    \textbf{Model} & \textbf{\# MFCC} & \textbf{Original shape} & \textbf{Shape after PCA} &
    \textbf{TEV} & \textbf{C} & \textbf{gamma} & \textbf{kernel} & \textbf{Grid time (s)} \\
    \midrule

    SVM & 40 & (5845, 40) & (5845, 30) & 0.97 & 10   & {auto}  & rbf & 24.58 \\
    SVM & 50 & (5845, 50) & (5845, 30) & 0.94 & 10   & {scale} & rbf & 23.60 \\
    SVM & 60 & (5845, 60) & (5845, 30) & 0.91 & 1    & {auto}  & rbf & 26.63 \\

    \bottomrule
  \end{tabularx}

  \vspace{3pt}
  \raggedright\footnotesize\textit{Notes:} TEV = total explained variance after PCA. `gamma` values such as `auto`/`scale` are shown verbatim; numeric values (if used) will align via `siunitx`. Grid times are wall-clock seconds measured on the tuning workstation.
\end{table*}

\subsection{Random Forest parameter optimisation}

We performed a grid search to identify robust Random Forest (RF) hyperparameters using \gls{mfcc}-derived inputs preprocessed by PCA. PCA truncated each input set to 30 components while retaining the majority of variance (high TEV). Table~\ref{tab:rf_arc} summarises the best hyperparameters discovered for each initial \gls{mfcc} count.

For each configuration we report TEV after PCA and the RF hyperparameters that maximised validation performance: tree depth (`max\_depth`), minimum samples per leaf and split, number of estimators, and the wall-clock grid-search time. These results reveal the degree of tree complexity and minimal leaf-size regularisation required under the different input dimensionalities.

\begin{table*}[htbp]
  \centering
  \scriptsize
  \caption{Random Forest grid-search and PCA settings. PCA reduced features to 30 components while preserving most variance.}
  \label{tab:rf_arc}
  \setlength{\tabcolsep}{4pt}
  \begin{tabularx}{\textwidth}{@{}
      l 
      c 
      p{1.8cm} 
      p{1.8cm} 
      S[table-format=1.2] 
      p{1.4cm} 
      c 
      c 
      S[table-format=3.0] 
      S[table-format=5.2] 
      @{}}
    \toprule
    \textbf{Model} & \textbf{\# MFCC} & \textbf{Original shape} & \textbf{Shape after PCA} &
    \textbf{TEV} & \textbf{max\_depth} & \textbf{min\_samples\_leaf} & \textbf{min\_samples\_split} & \textbf{n\_estimators} & \textbf{Grid time (s)} \\
    \midrule

    RF & 40 & (5845, 40) & (5845, 30) & 0.97 & {None} & 1 & 2  & 200 & 61.36 \\
    RF & 50 & (5845, 50) & (5845, 30) & 0.94 & 20     & 3 & 2  & 200 & 61.78 \\
    RF & 60 & (5845, 60) & (5845, 30) & 0.91 & {None} & 1 & 10 & 200 & 62.56 \\

    \bottomrule
  \end{tabularx}

  \vspace{3pt}
  \raggedright\footnotesize\textit{Notes:} TEV = total explained variance after PCA. `None` indicates unrestricted tree depth. `n\_estimators` denotes the number of trees in the forest. Grid times are wall-clock seconds on the tuning machine.
\end{table*}

\subsection{Naive Bayes parameter optimisation}

We performed a focused grid search to determine the Naive Bayes (NB) configuration that yields robust classification performance on \gls{mfcc}-derived feature sets preprocessed with principal component analysis (\gls{pca}). For these experiments PCA was applied to decorrelate the features and then truncated to a fixed 30 components—this choice reflects consistently high total explained variance (TEV) across the tested initial feature counts. 

Table~\ref{tab:nb_arc} summarises the optimisation results for the different initial \gls{mfcc} dimensions. For each configuration we report TEV after PCA, the number of PCA components retained, mean and standard deviation of the test score (cross-validation), the ranking of the best configuration, and wall-clock grid-search time. The results indicate stable and high predictive performance across input dimensionalities, with minimal tuning cost and low variance in validation performance, supporting the choice of 30 PCA components as a parsimonious and effective representation for NB in this task.

\begin{table*}[htbp]
  \centering
  \scriptsize
  \caption{\gls{nb} grid-search and PCA settings. PCA was used to decorrelate features and a fixed 30 components were retained for subsequent classification.}
  \label{tab:nb_arc}
  \setlength{\tabcolsep}{6pt}
  \begin{tabularx}{\textwidth}{@{} 
      l  
      c  
      p{1.6cm} 
      p{1.6cm} 
      S[table-format=1.2] 
      S[table-format=2.0] 
      S[table-format=1.6] 
      S[table-format=1.6] 
      c  
      S[table-format=4.2] 
      @{}}
    \toprule
    \textbf{Model} & \textbf{\# MFCC} & \textbf{Original shape} & \textbf{Shape after PCA} &
    \textbf{TEV} & \textbf{\# PCA comps} & \textbf{Mean test score} & \textbf{Std test score} & \textbf{Rank} & \textbf{Grid time (s)} \\
    \midrule

    Naive Bayes & 40 & (5845, 40) & (5845, 30) & 0.97 & 30 & 0.929600 & 0.011571 & 1 & 0.28 \\
    Naive Bayes & 50 & (5845, 50) & (5845, 30) & 0.94 & 30 & 0.935672 & 0.010552 & 1 & 0.81 \\
    Naive Bayes & 60 & (5845, 60) & (5845, 30) & 0.91 & 30 & 0.933105 & 0.005837 & 1 & 0.56 \\

    \bottomrule
  \end{tabularx}

  \vspace{3pt}
  \raggedright\footnotesize\textit{Notes:} TEV = total explained variance after PCA. Mean and standard deviation refer to the cross-validation test score (reported here to four decimal places). Grid-search times are wall-clock seconds measured on the tuning workstation.
\end{table*}

\subsection{XGBoost parameter optimisation}

We conducted a grid search to identify the optimal \gls{xgboost} hyperparameters for classification using \gls{mfcc} features preprocessed with principal component analysis (\gls{pca}). PCA was used primarily to decorrelate features while preserving the original dimensionality when total explained variance (TEV) remained high. 

Table~\ref{tab:xgboost_arc} summarises the best-performing hyperparameter configurations for different initial \gls{mfcc} counts. For each configuration we report TEV after PCA and the hyperparameters that produced the best validation performance: column subsampling (`colsample\_bytree`) and row subsampling (`subsample`) (regularisation), learning rate, maximum tree depth (`max\_depth`) (model complexity), and the number of estimators. We also report wall-clock grid-search time to give an empirical sense of tuning cost. These results show modest shifts in regularisation and learning-rate settings as the input dimensionality changes, while the number of estimators remained consistent across experiments.

\begin{table*}[htbp]
  \centering
  \scriptsize
  \caption{\gls{xgboost} grid-search and PCA settings. PCA was applied to decorrelate features while preserving original dimensionality when TEV remained high.}
  \label{tab:xgboost_arc}
  \setlength{\tabcolsep}{3pt}
  \begin{tabularx}{\textwidth}{@{} 
      l   
      c   
      p{1.6cm} 
      p{1.6cm} 
      S[table-format=1.2] 
      S[table-format=1.2] 
      S[table-format=1.2] 
      c   
      S[table-format=3.0] 
      S[table-format=1.2] 
      S[table-format=6.2] 
      @{}}
    \toprule
    \textbf{Model} & \textbf{\# MFCC} & \textbf{Original shape} & \textbf{Shape after PCA} &
    \textbf{TEV} & \textbf{colsample\_bytree} & \textbf{learning\_rate} & \textbf{max\_depth} & \textbf{n\_estimators} & \textbf{subsample} & \textbf{Grid time (s)} \\
    \midrule

    XGBoost & 40 & (5845, 40) & (5845, 30) & 0.97 & 0.70 & 0.10 & 5 & 300 & 0.70 & 28.16 \\
    XGBoost & 50 & (5845, 50) & (5845, 30) & 0.94 & 0.90 & 0.20 & 4 & 300 & 0.80 & 27.79 \\
    XGBoost & 60 & (5845, 60) & (5845, 30) & 0.91 & 0.90 & 0.10 & 4 & 300 & 0.90 & 24.73 \\

    \bottomrule
  \end{tabularx}

  \vspace{3pt}
  \raggedright\footnotesize\textit{Notes:} TEV = total explained variance after PCA. `colsample\_bytree` and `subsample` control feature- and row-level regularisation respectively; `n\_estimators` denotes the number of boosted trees. Grid times are wall-clock seconds measured on the tuning machine.
\end{table*}

\subsection{MLP parameter optimisation}

We performed a systematic grid search to optimise the multilayer perceptron (MLP) architecture for classification using \gls{mfcc} features. Principal component analysis (\gls{pca}) was applied primarily to decorrelate features; in this study we retained the original feature dimensionality (i.e., PCA served as a rotation rather than a dimensionality reduction) because the cumulative total explained variance (\gls{tev}) remained high. 

Table~\ref{tab:mlp_arc} summarises the grid-search results for different initial \gls{mfcc} feature counts. For each configuration we report the total explained variance after PCA, the hyperparameters that produced the best validation performance (activation, hidden-layer size, learning-rate policy, solver, and regularisation strength $\alpha$), and the wall-clock time required for the grid search. The selected hyperparameters reflect the trade-offs between representational capacity and regularisation required for our dataset; runtimes provide an indication of the computational cost of tuning.

\begin{table*}[htbp]
  \centering
  \scriptsize
  \caption{MLP grid-search and PCA settings (PCA used for decorrelation while retaining original dimensionality).}
  \label{tab:mlp_arc}
  \setlength{\tabcolsep}{6pt}
  \begin{tabularx}{\textwidth}{@{} 
      l  
      c  
      p{1.8cm} 
      p{1.8cm} 
      S[table-format=1.2] 
      c  
      c  
      c  
      c  
      S[table-format=1.4] 
      S[table-format=6.2] 
      @{}}
    \toprule
    Model & \# MFCC & Original shape & Shape after PCA & {TEV} & Activation & Hidden & {LR policy} & Solver & {alpha} & {Grid time (s)} \\
    \midrule
    MLP & 40 & (5845,40) & (5845,40) & 0.97 & tanh & 128 & constant & adam & 0.0010 & 110.14 \\
    MLP & 50 & (5845,50) & (5845,50) & 0.98 & relu & 64  & constant & adam & 0.0100 & 90.02  \\
    MLP & 60 & (5845,60) & (5845,60) & 0.99 & relu & 64  & constant & adam & 0.0001 & 80.06  \\
    \bottomrule
  \end{tabularx}
  \vspace{2pt}

  \raggedright\footnotesize\textit{Notes:} Shapes shown as (samples, features). TEV = total explained variance after PCA.
\end{table*}


\subsection{1D-CNN parameter optimisation with one hidden layer}
Table~\ref{tab:1dcnn1l_arc} reports the best hyperparameter configurations found for a \gls*{1d-cnn} with a single hidden layer, when the original \gls*{mfcc} feature sets (40, 50, 60 coefficients) are reduced to 30 components by PCA. For each setting we report the post-PCA feature shape, the total explained variance (TEV), the best-performing hyperparameters discovered by grid search (batch size, epochs, dense units, filters, kernel size, pooling) and elapsed grid-search time.

Across the explored configurations the optimisation consistently selected the same batch size, epoch count and dense-layer width, indicating a robust, compact architecture for this task. Kernel size and optimiser choices exhibited minor variation between MFCC settings, reflecting subtle dependencies of receptive-field design and optimisation dynamics on input dimensionality. Grid-search times (\(\approx\)30–36\,s) were short, indicating that single-hidden-layer configurations can be tuned efficiently; this makes them attractive for rapid prototyping and for deployment scenarios where compute budgets are limited.

\begin{table}[h!]
  \centering
  \caption{\gls{1d-cnn} (one hidden layer): PCA + grid-search optimisation results (vertical-rule style).}
  \label{tab:1dcnn1l_arc}
  \setlength{\tabcolsep}{4pt} 
  \renewcommand{\arraystretch}{1.05} 

  \begingroup
  \fontfamily{<your-font-here>}\selectfont
  \resizebox{\linewidth}{!}{%
    \begin{tabular}{|l|c|c|c|c|c|c|c|c|c|c|c|c|}
      \hline
      \textbf{Model} & \textbf{\#MFCCs} & \textbf{Orig.\ shape} & \textbf{PCA shape} & \textbf{TEV} &
      \textbf{Batch} & \textbf{Epochs} & \textbf{Dense} & \textbf{Filters} & \textbf{Kernel} &
      \textbf{Pool} & \textbf{Optimiser} & \textbf{Grid time (s)} \\ \hline

      1D-CNN (1 HL) & 40 & (5845, 40) & (5845, 30) & 0.97 & 32 & 10 & 128 & 64 & 5 & 2 & adam    & 35.51 \\ \hline
      1D-CNN (1 HL) & 50 & (5845, 50) & (5845, 30) & 0.94 & 32 & 10 & 128 & 64 & 3 & 2 & rmsprop & 35.65 \\ \hline
      1D-CNN (1 HL) & 60 & (5845, 60) & (5845, 30) & 0.91 & 32 & 10 & 128 & 64 & 5 & 2 & rmsprop & 30.12 \\ \hline

    \end{tabular}%
  }
  \endgroup

\end{table}

\paragraph{Recommendations.}
For reproducible and efficient hyperparameter search we recommend:
\begin{itemize}
  \setlength\itemsep{0pt}
  \item using randomized or Bayesian optimisers (e.g., Optuna, Hyperopt) to explore the space more efficiently than exhaustive grid search;
  \item adopting early stopping and short pilot runs to prune poor configurations quickly;
  \item employing multi-fidelity tuning (subsampled data, reduced epochs) prior to full training;
  \item logging experiments (MLflow, Weights \& Biases or DVC) to ensure reproducibility and to enable warm-starting of subsequent searches.
\end{itemize}

These practices retain or improve performance while substantially cutting wall-clock time and compute cost, which is particularly important for lifecycle management and operational deployment.

\subsection{Parameter optimisation for 1D-CNN with two hidden layers}
Table~\ref{tab:1dcnn2l_arc} summarises the optimal architectures and training hyperparameters found for the \gls{1d-cnn} with two hidden layers when varying the number of \gls{mfcc} features and applying PCA for dimensionality reduction. The table reports the post-PCA feature shape, total explained variance, best-performing hyperparameters discovered by grid search (batch size, epochs, dense units, filters, kernel sizes, pooling sizes and optimiser), and the elapsed grid-search time in seconds.

The results indicate consistent high performance across settings but also highlight that exhaustive grid search for \gls{dl} architectures is computationally expensive: the reported grid-search times (\(\approx\)11,000–11,200\,s) reflect substantial compute. For practical model development we recommend complementary strategies to reduce computation while preserving search effectiveness, such as randomized or Bayesian hyperparameter search, early stopping, progressive resizing, lower-fidelity estimates (reduced epochs or subset of data), and transfer learning with fine-tuning. These approaches can dramatically reduce wall-clock time and cost while finding comparably good hyperparameters.

\begin{table}[htbp]
  \centering
  \small
  \caption{\gls{1d-cnn} (two hidden layers): PCA + grid-search optimisation results.}
  \label{tab:1dcnn2l_arc}
  \setlength{\tabcolsep}{8pt}
  \resizebox{\textwidth}{!}{%
  \begin{tabular}{|l|c|c|c|c|c|c|c|c|c|}
    \hline
    \textbf{Model} &
    \textbf{Feature shape (post-PCA)} &
    \textbf{Total Expl.\ Var.} &
    \textbf{Batch} &
    \textbf{Epochs} &
    \textbf{Dense} &
    \textbf{Filters (L1,L2)} &
    \textbf{Kernel (L1,L2)} &
    \textbf{Pool (L1,L2)} &
    \textbf{Grid time (s)} \\ \hline\hline

    1D-CNN (2 HL) & (5845, 30) & 0.97 & 32  & 20 & 64  & 64, 128  & 5, 5 & 2, 2 & 11114.43 \\ \hline
    1D-CNN (2 HL) & (5845, 30) & 0.94 & 64  & 20 & 128 & 128, 256 & 3, 5 & 2, 2 & 11112.93 \\ \hline
    1D-CNN (2 HL) & (5845, 30) & 0.91 & 64  & 20 & 256 & 64, 128  & 5, 5 & 2, 2 & 11223.71 \\ \hline
  \end{tabular}%
  }
\end{table}

Given the heavy computational burden of exhaustive grid searches for \gls{1d-cnn} hyperparameters, recommended approaches entail (1) using randomized or Bayesian optimization, such as Optuna, Hyperopt, instead of full-grid search to explore hyperparameter space more efficiently, (2) employing early stopping and reduced-epoch pilot runs to prune poor configurations quickly, (3) adopting multi-fidelity optimization (shorter runs or smaller data subsets) before committing to full training, (4) leveraging transfer learning or pre-trained backbones where applicable, followed by light fine-tuning, (5) logging and versioning experiments like MLflow, Weights \& Biases, or DVC,  to ensure reproducibility and enable warm-starting of searches. These actions typically reduce search time substantially while retaining or improving model performance and reproducibility.

\subsection{1D-CNN Architecture}
A \gls*{1d-cnn} is a neural network architecture tailored to one-dimensional sequential data such as time series or audio. A canonical design begins with a 1D input layer and proceeds through a sequence of convolutional blocks: small-kernel 1D convolutional layers (typically with ReLU activation) interleaved with pooling layers for progressive feature aggregation and dimensionality reduction. After the convolutional stack, a Flatten (or global-pooling) layer converts feature maps to a vector that feeds one or more fully connected (Dense) layers with optional Dropout for regularization, finishing with a task-dependent output layer (e.g., sigmoid for binary classification or softmax for multi-class).

\begin{figure}[h!]
\centering
\includegraphics[width=0.7\textwidth]{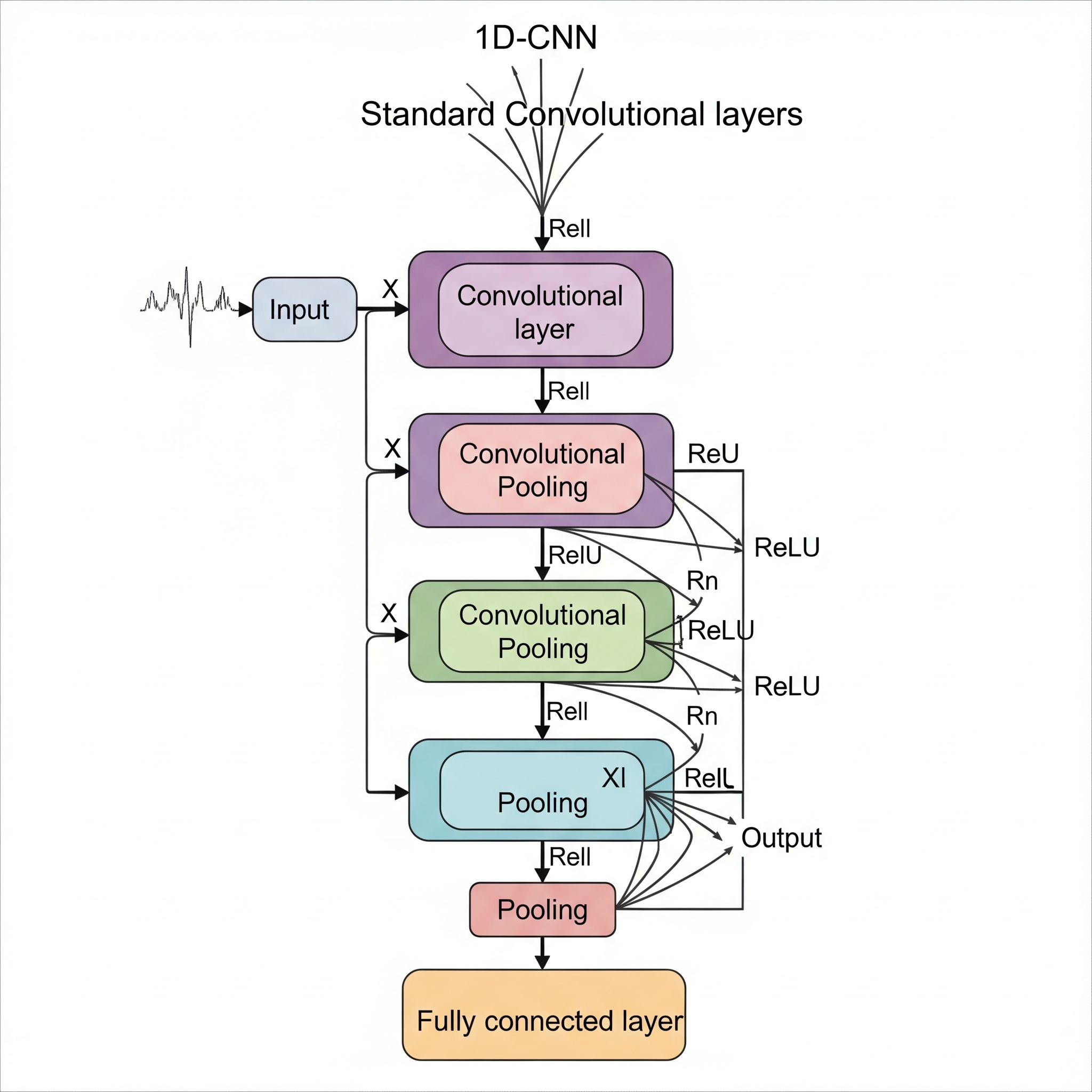}
\caption{Generic schematic of the \gls*{1d-cnn} architecture used in this study.}
\label{fig:1d-cnn_architecture}
\end{figure}

In practice, design choices such as kernel size, number of filters per layer, pooling strategy, and the depth of the network determine the trade-off between representational capacity and inference cost. Small kernels (e.g., 3–7 samples) stacked across multiple layers allow the network to learn hierarchical temporal features, while pooling (max or average) controls temporal resolution and reduces compute. For resource-constrained deployments, shallow architectures with fewer filters and modest pooling often provide a favourable balance of accuracy and latency; for research experiments, modest increases in depth and filter count can improve performance but incur higher training and inference cost. Hyperparameters should be selected via cross-validation and evaluated in terms of both predictive metrics and operational constraints (latency, memory, power).

\subsection{1D-DCNN Architecture}
A \gls{1d-dcnn} extends the \gls{1d-cnn} paradigm by using dilated convolutions to enlarge the receptive field without substantially increasing parameter count or reducing temporal resolution. The typical architecture begins with a 1D input layer for sequential data, followed by multiple blocks of dilated 1D convolutional layers (usually with ReLU activation), optionally interleaved with pooling. Dilated convolutions apply filters with a specified \emph{dilation rate}, effectively inserting gaps between filter taps and enabling the network to model long-range dependencies while keeping networks relatively shallow. After the convolutional blocks, a Flatten (or global pooling) layer prepares features for fully connected (Dense) layers with optional Dropout for regularization, concluding with a task-specific output layer (e.g., sigmoid for binary, softmax for multi-class).

\begin{figure}[h!]
\centering
\includegraphics[width=0.7\textwidth]{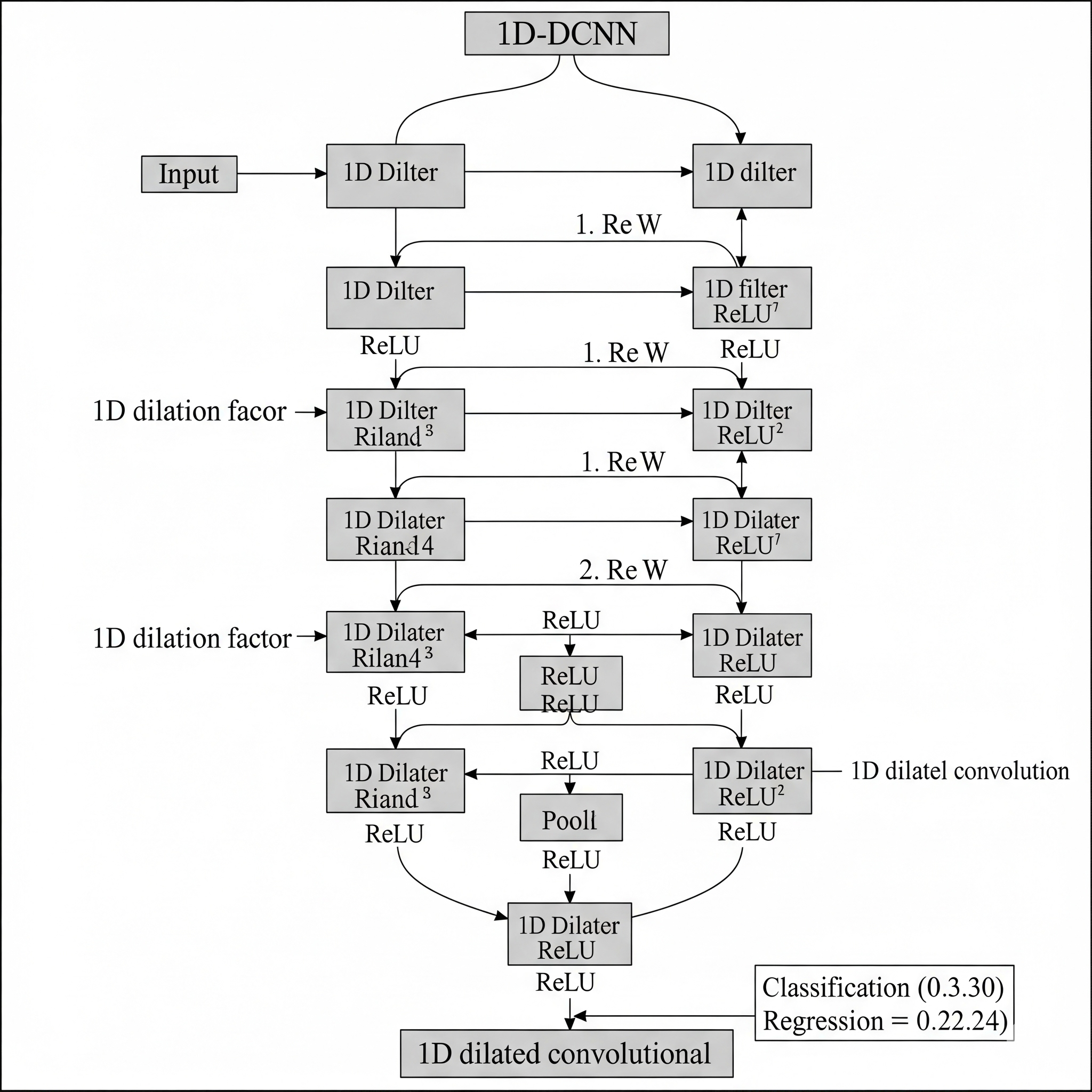}
\caption{Schematic of the \gls*{1d-dcnn} architecture used in this study.}
\label{fig:1dcnn_architecture}
\end{figure}

Compared with a standard \gls{1d-cnn}, the principal distinction is the convolution operator. Standard 1D convolutions process contiguous input segments, while dilated convolutions insert defined gaps (dilation) between kernel elements, permitting a much larger effective receptive field for the same kernel size and depth. Practically, this yields three main advantages for long sequential signals such as audio: (1) the ability to capture long-range context without deep stacking of layers, (2) improved parameter efficiency when modeling extended dependencies, and (3) preservation of temporal resolution because dilated convolutions can reduce the need for aggressive downsampling. These properties make \gls{1d-dcnn} particularly suitable for tasks where distant temporal structure is informative but excessive pooling would lose fine-grained cues.

There are trade-offs: dilation patterns must be chosen carefully (e.g., exponential vs linear schedules) to avoid blind spots in coverage, and dilated layers may be more sensitive to hyperparameter settings (kernel size, dilation rates, and regularization). In practice, shallow dilated stacks (one or two dilated blocks) often provide a good balance of performance and inference cost for our lobster audio tasks; further gains are attainable through systematic tuning, judicious use of global pooling, and model-compression techniques for deployment on resource-constrained hardware.

\section{Discussion}\label{sec:discussion}
\gls{ml} and \gls{dl} approaches exhibited complementary strengths on the sex and age bioacoustic tasks.  Binary classification of adult versus juvenile achieved the strongest results overall (see Tables~\ref{tab:ml_adult_juvenile_metrics} and~\ref{tab:dl_adult_juvenile_metrics}): the weakest performer was \gls{nb} (89.12\%), while the best single model was \gls{svm} (98.50\%). For male versus female classification the lowest accuracy was also from \gls{nb} (81.60\%), and \gls{svm} again attained the highest score (96.17\%). A \gls{1d-dcnn} with two hidden layers achieved the second-highest age-class accuracy (98.08\%), but the difference between 98.50\% and 98.08\% is not large enough to conclude a clear superiority given the experimental scope and sample counts. Notably, \gls{svm} showed greater stability as the number of \gls{mfcc} coefficients varied (98.29\%, 98.50\%, 98.29\% for 40/50/60 coeffs), whereas the \gls{1d-dcnn} (2 hidden layers) followed 98.08\%, 97.74\%, 97.47\% for the same settings. Because only three MFCC settings were explored, these trends are suggestive but not definitive; a denser sweep and graphical analysis of accuracy versus MFCC dimensionality would be required to establish a functional relationship.

Across models, \gls{svm} delivered the best balance of accuracy and consistency for the Adult vs Juvenile task (peak accuracy 98.50\% and F1 score 99.76\%). \gls{knn} provided the fastest inference time (0.87\,ms) with competitive accuracy (98.08\%). \gls{xgboost} and \gls{mlp} produced strong results (\(\approx\)97–98\% accuracy; F1 \(\approx\)97–98\%) with inference times typically lower than \gls{svm} in our tests. \gls{rf} performed well (97.26\% accuracy; F1 \(\approx\)95.79\%), while \gls{nb} consistently performed worst across metrics despite a low inference time (2.10\,ms). \gls{dl} variants (1D-CNN and the 1D-DCNN family) attained comparable classification accuracy (generally $>$97\%) and very high \gls{auc-roc} values ($>$99.5\%), but at substantially higher inference times than traditional \gls{ml} models. Increasing the number of hidden layers in \gls{dl} models tended to increase inference time without reliably improving accuracy; shallow architectures (one or two hidden layers) therefore appear preferable given our data and computational constraints.

For practical, real-time deployments (e.g., \gls{pam} or underwater \gls{uiot} edge systems \cite{mackridge2018practical}), model selection must trade predictive performance against latency, power and memory budgets. In this context, compact \gls{ml} pipelines built on efficient feature extractors (such as \gls{mfcc}) and lightweight classifiers (e.g., \gls{svm}, \gls{knn}, or modest tree ensembles) often provide the best operational compromise; \gls{dl} models can match or slightly exceed accuracy but require model-compression and efficient-inference engineering to be viable on resource-constrained hardware \cite{han2015,sze2017}. The optimal architecture will therefore depend on the specific deployment constraints (sampling cadence, available compute, acceptable latency) and on whether probabilistic calibration or marginal improvements in accuracy are prioritized.

Observed performance differences between age and sex classifications likely reflect dataset factors (size, class imbalance, recording quality) and the relative strength of acoustic cues: ontogenetic or age-related signals appear more salient than sex-specific cues in our tank recordings. Addressing these limitations—by increasing sample sizes, improving annotation quality, and expanding in-situ (natural habitat) recordings—should improve model generalization and clarify the roles of feature dimensionality and architecture depth. Future work should also evaluate sequence models and transformer-style architectures, carry out systematic hyperparameter and feature-selection studies, and perform rigorous cross-region validation to assess transferability \cite{gong2021,verma2021}.

Finally, it is important to situate these technical advances within the broader socio-ecological context. EU fisheries management has a mixed record with respect to overfishing and enforcement, and lobster stocks face region-specific depletion risks, size-structure concerns and climate-driven shifts in distribution and productivity \cite{pew2021,eca2022,cefas2023,matic2022}. \gls{ai}-enhanced monitoring can provide high-resolution observational inputs to support management, but technological solutions are complementary tools rather than substitutes for robust governance, compliance and ecosystem-based management. Integrating AI-derived indicators into policy requires transparent methods, reproducible data and explicit attention to the social and institutional dimensions of fisheries governance.

\section{Conclusions}\label{sec:conclusion}
In conclusion, while \gls{dl} models can deliver very high accuracy for lobster sound classification, their substantial training and inference costs — and the engineering required for edge deployment — often make traditional \gls{ml} approaches (e.g., \gls{svm}, \gls{knn}, \gls{xgboost}) the more pragmatic choice for many real-world bioacoustic applications \cite{strubell2019,sze2017,chen2016}. Efficient feature extractors such as \gls{mfcc} remain powerful front-ends; when paired with compact classifiers they frequently achieve an attractive trade-off between predictive performance and computational cost \cite{davis1980,han2015,strubell2019}.

Our results demonstrate that \gls{Hom_euro} rasping and clicking are classifiable with \gls{ai} models and that both conventional \gls{ml} and \gls{dl} approaches are viable for age and sex discrimination under controlled (tank) conditions. For operational use within \gls{pam} and underwater \gls{uiot} systems, model selection must balance discrimination against latency, power and memory constraints; model-compression and efficient-inference techniques can mitigate but not eliminate the overhead of large \gls{dl} architectures \cite{han2015,sze2017}. Observed performance differences between adults vs juveniles and males vs females likely reflect dataset limitations (size, imbalance, recording quality) and weaker sex-specific acoustic cues; expanding representative sampling and improving in-situ annotation will be important next steps. Future work will explore sequence models and transformer-style architectures and perform rigorous cross-region validation to assess transferability \cite{gong2021,verma2021}.

Implementing the proposed stacking-ensemble pipeline would bring several concrete benefits. By combining diverse base learners (trees, kernel methods, and neural networks) into a stacked meta-learner, the pipeline typically improves discrimination and reduces variance relative to single models \cite{wolpert1992,breiman1996stacked,zhou2012ensemble}. Boosting and bagging-based components (e.g., \gls{xgboost}, \gls{rf}) capture complementary signal structure in noisy bioacoustic data and tend to generalize more robustly across heterogeneous conditions \cite{chen2016,dietterich2000ensemble}. The OOF-based stacking design prevents information leakage and enhances reproducibility, while a regularized meta-learner often yields better-calibrated probabilistic outputs — a critical asset when model outputs feed management thresholds. These gains come with practical trade-offs: fold-wise training and CNN retraining increase compute and require disciplined evaluation, model registry and monitoring (drift detection, scheduled re-training). In our judgment, the modest operational overhead is justified by the improvements in accuracy, robustness and calibrated uncertainty that support management decisions.

Finally, to enhance transparency and reproducibility we have published the data and code with ISO-compliant metadata and included ensemble baselines to strengthen predictive comparisons. Together, these measures advance the practical applicability of \gls*{ai}-enhanced bioacoustics for lobster management and conservation.

\section{Acknowledgements}\label{sec:acknowledge}
We are profoundly grateful to Murray McBay and Co., particularly their Lobster Shop in Scotland, for their generous support in allowing us to record lobster sounds within their facility. Their invaluable assistance in sorting female, male, juvenile, and adult lobsters was crucial to this research. We extend our sincere thanks to Dr. Nicola Khan for her pivotal role in establishing the connection with Murray McBay and Co. \par
The work reported in this article contributes to the EnviroBrain Impact Case Study.

\bibliographystyle{elsarticle-harv} 

\bibliography{references}\label{sec:ref}

\appendix \label{sec:appendix}

\section{Specifications of the Raspberry Pi 3 Model B}
Table~\ref{tab:rpi3specs} provides a comprehensive technical overview of the Raspberry Pi 3 Model B. It details various hardware and software aspects, categorized into sections like CPU, RAM, Ports (covering video, audio, USB, Ethernet, wireless, Bluetooth, GPIO, camera, display, storage, and power), and Software (listing the primary OS, desktop environment, included software, and other compatible operating systems). Essentially, it serves as a detailed spec sheet for the Raspberry Pi 3 Model B. 

\begin{table}[h!]
\centering
\caption{Specifications of the Raspberry Pi 3 Model B}
\label{tab:rpi3specs}
\begin{tabular}{|p{4cm}|p{9cm}|} 
\hline
\textbf{Category} & \textbf{Specification} \\
\hline
\textbf{CPU} & \\
\quad SoC & Broadcom BCM2837 \\
\quad Architecture & 64-bit ARM Cortex-A53 \\
\quad Core Count & Quad-core (4 cores) \\
\quad Clock Speed & 1.2 GHz \\
\hline
\textbf{RAM} & \\
\quad Size & 1GB LPDDR2 SDRAM \\
\hline
\textbf{Ports} & \\
\quad Video Output & Full-size HDMI (up to 1080p60), Composite (3.5mm TRRS) \\
\quad Audio Output & Analog (3.5mm TRRS), Digital (HDMI) \\
\quad USB Ports & 4 x USB 2.0 \\
\quad Ethernet & 10/100 BaseT RJ45 \\
\quad Wireless & 802.11 b/g/n Wi-Fi (2.4 GHz) \\
\quad Bluetooth & Bluetooth 4.1 Classic, Bluetooth Low Energy (BLE) \\
\quad GPIO & 40-pin header (26 GPIO, Power, Ground) \\
\quad Camera Interface (CSI) & 15-pin MIPI CSI \\
\quad Display Interface (DSI) & 15-pin MIPI DSI \\
\quad Storage & MicroSD Card Slot \\
\quad Power Input & Micro USB (5V/2.5A recommended) \\
\hline
\textbf{Software (Commonly Equipped With)} & \\
\quad Primary OS & Raspberry Pi OS (Debian-based Linux) \\
\quad Desktop Environment & LXDE \\
\quad Included Software & Chromium Browser, LibreOffice (optional), Python, Scratch, etc. \\
\quad Other Compatible OS & Ubuntu MATE/Server, Windows 10 IoT Core, LibreELEC, RetroPie, RISC OS, etc. \\
\hline
\end{tabular}
\end{table}

\section{Hardware and Software resources}
Table~\ref{tab:resources} provides a concise overview of the key resources (both hardware and software) utilized in a project involving dataset collection and processing. It lists each resource, such as Raspberry Pi 3, Hydrophone, Python 3.12.3, PyAudio, Laptop (PC), ThinkPad, Keras, Scikit-learn, and TensorFlow, alongside a brief description of its role or function in the project. The table essentially acts as a quick reference guide to the technological stack employed. The details of the computer used is Table~\ref{tab:thinkpad_pf2d95wh}

\begin{table}[h!]
\centering
\caption{Hardware and Software resources for dataset collection and processing} 
\label{tab:resources}
\begin{tabular}{|l|p{8cm}|} 
\hline
\textbf{Resources} & \textbf{Description} \\
\hline
Raspberry Pi 3 & Hardware and software equipped with CPU, Ports, RAM\\
\hline
Hydrophone & Hardware, underwater audio sensor \\
\hline
Python 3.12.3 & Software, development environment \\
\hline
PyAudio & Software, library resource for recording sound. \\
\hline
Laptop (PC), ThinkPad & Housing Python, SSH, networking capabilities. \\
\hline
Keras & High-level neural networks API written in Python \\
\hline
Scikit-learn & Python library for \gls{ml}, built on top of NumPy, SciPy, and Matplotlib, offer some basic tools for building and training deep neural networks \\
\hline
TensorFlow & An open-source \gls{ml} framework developed by Google, designed to build and train various types of neural networks \\
\hline
\end{tabular}
\end{table}

\begin{table}[h]
\centering
\caption{Technical specifications of the Lenovo ThinkPad X1 Carbon Gen 13 Aura Edition (PF-2D95WH).}
\label{tab:thinkpad_pf2d95wh}
\begin{tabular}{p{0.32\linewidth} p{0.62\linewidth}}
\hline
\textbf{Category} & \textbf{Specification} \\
\hline
\textbf{Performance} & Intel Core Ultra (Series 2) processor on Intel vPro, Evo Edition platform; integrated AI engine with up to 13~TOPS AI performance; optimized performance per watt for extended battery life. \\
\textbf{AI functions} & Local AI features including enhanced video conferencing, document automation, email management, scheduling, improved responsiveness, and enhanced data privacy. \\
\textbf{Display} & 14-inch OLED, 2.8K (2880~$\times$~1800) resolution, 120~Hz refresh rate, 400--500~nits brightness, 100\% DCI-P3 color gamut, DisplayHDR True Black 500 certified. \\
\textbf{Memory and storage} & Up to 64~GB soldered LPDDR5x (8400~MT/s) RAM; up to 2~TB PCIe Gen5 SSD. \\
\textbf{Design and construction} & Lightweight design $<$1~kg; chassis built from recycled carbon fiber, magnesium, and plastic; MIL-STD-810H certified durability. \\
\textbf{Security} & ThinkShield suite: discrete TPM (dTPM), biometric authentication, AI-enhanced threat detection. \\
\textbf{Usability} & Haptic TouchPad; TrackPoint with double-tap menu for audio/video settings; Dictation Toolbar with speech-to-text support. \\
\textbf{Connectivity} & USB-C Thunderbolt 4, USB-A (full-size), and other modern I/O options. \\
\textbf{Power} & Rapid Charge technology; compact GaN charger for fast charging. \\
\hline
\end{tabular}
\end{table}


\section{Hardware: Hydrophone Features}
Table~\ref{tab:hydrophone_features} provides a detailed breakdown of the key characteristics and capabilities of a specific hydrophone device. 
\begin{table}[htbp]
  \centering
  \caption{Hydrophone specification — Aquarian Audio Model H2d (3.5\,mm cable termination)}
  \label{tab:hydrophone_features}
  \small
  \setlength{\tabcolsep}{8pt}
  \renewcommand{\arraystretch}{1.2}
  \begin{tabular}{|p{5.0cm}|p{8.5cm}|}
    \hline
    \textbf{Specification} & \textbf{Model H2d (Aquarian Audio)} \\
    \hline\hline
    Manufacturer & Aquarian Audio \\
    \hline
    Model & H2d \\
    \hline
    Termination & 3.5\,mm cable termination (note: termination only—electrical/acoustic specifications identical to other termination options) \\
    \hline
    Sensitivity & -172\,dB re: 1\,V/$\mu$Pa (±4\,dB, 20\,Hz–4\,kHz). Approx.\ sensitivity at 100\,kHz: -210\,dB re: 1\,V/$\mu$Pa. \\
    \hline
    Useful range & $<$10\,Hz to $>$100\,kHz (not measured above 100\,kHz). \\
    \hline
    Polar response & Omnidirectional (horizontal). \\
    \hline
    Operating depth & $<$80\,m. \\
    \hline
    Output impedance & 1.1\,k$\Omega$ (typical). \\
    \hline
    Power (current) & 0.7\,mA (typical). \\
    \hline
    Minimum PIP requirement & 2\,V, 0.7\,mA. \\
    \hline
    Dimensions & 25\,mm $\times$ 46\,mm. \\
    \hline
    Mass & 105\,g. \\
    \hline
    Specific gravity & 5.3. \\
    \hline
  \end{tabular}

  \vspace{4pt}
  \footnotesize{\textbf{Note:} The “3.5\,mm cable termination” refers only to the connector termination. All acoustic and electrical specifications in the table apply to the H2d hydrophone irrespective of connector type unless otherwise specified. Replace any placeholder numbers if you have manufacturer updates.}
\end{table}


\end{document}